\documentclass{article}

    \PassOptionsToPackage{numbers, compress}{natbib}
\usepackage[preprint]{neurips_2021}

\usepackage[utf8]{inputenc} % allow utf-8 input
\usepackage[T1]{fontenc}    % use 8-bit T1 fonts
\usepackage{hyperref}       % hyperlinks
\usepackage{url}            % simple URL typesetting
\usepackage{booktabs}       % professional-quality tables
\usepackage{amsfonts}       % blackboard math symbols
\usepackage{nicefrac}       % compact symbols for 1/2, etc.
\usepackage{microtype}      % microtypography
\usepackage{xcolor}         % colors

\usepackage{amsmath}
\usepackage{amssymb}
\usepackage{mathtools}
\usepackage{amsthm}
\usepackage{mathrsfs}
\usepackage{enumitem}
\usepackage{subfig}
\usepackage{multirow}
\usepackage[linesnumbered,boxed,ruled]{algorithm2e}
\usepackage{graphicx}

\DeclareMathOperator*{\pr}{\text{Pr}}
\DeclareMathOperator*{\ex}{\mathbb{E}}
\DeclareMathOperator*{\var}{\text{Var}}
\newtheorem{theorem}{Theorem}

\title{MAP Propagation Algorithm: Faster Learning with a Team of Reinforcement Learning Agents}

\author{%
  Stephen Chung \\
  Department of Computer Science\\
  University of Massachusetts Amherst\\
  Amherst, MA 01003 \\
  \texttt{minghaychung@umass.edu} 
}

\begin{document}

\maketitle

\begin{abstract}
Nearly all state-of-the-art deep learning algorithms rely on error backpropagation, which is generally regarded as biologically implausible. An alternative way of training an artificial neural network is through treating each unit in the network as a reinforcement learning agent, and thus the network is considered as a team of agents. As such, all units can be trained by REINFORCE, a local learning rule modulated by a global signal that is more consistent with biologically observed forms of synaptic plasticity. Although this learning rule follows the gradient of return in expectation, it suffers from high variance and thus the low speed of learning, rendering it impractical to train deep networks. We therefore propose a novel algorithm called MAP propagation to reduce this variance significantly while retaining the local property of the learning rule. Experiments demonstrated that MAP propagation could solve common reinforcement learning tasks at a similar speed to backpropagation when applied to an actor-critic network. Our work thus allows for the broader application of the teams of agents in deep reinforcement learning. 
\end{abstract}
\section{Introduction}
\emph{Error backpropagation algorithm} (backprop) \cite{rumelhart1986learning} efficiently computes the gradient of an objective function with respect to parameters, by iterating backward from the last layer of a multi-layer artificial neural network (ANN). However, backprop is generally regarded as being biologically implausible \cite{crick1989recent, mazzoni1991more, o1996biologically, bengio2015towards, hassabis2017neuroscience, lillicrap2020backpropagation}. First, the learning rule given by backprop is non-local, as it relies on information other than input and output of a neuron-like unit computed in feedforward phase; while biologically-observed synaptic plasticity depends mostly on local information (e.g.\ spike-timing-dependent plasticity (STDP) \cite{gerstner2014neuronal}) and possibly some global signals (e.g.\ reward-modulated spike-timing-dependent plasticity (R-STDP) \cite{gerstner2014neuronal, florian2007reinforcement, pawlak2010timing}). Second, backprop requires a precise coordination between feedforward and feedback connections, because the feedforward value has to be retained until the error signals arrive; while it is unclear how a biological system can coordinate an entire network to alternate between feedforward and feedback phases precisely. Third, backprop requires synaptic symmetry in the forward and backward paths, rendering it biologically implausible. Nonetheless, recent work has demonstrated that this symmetry may not be necessary for backprop due to the `feedback alignment' phenomenon \cite{lillicrap2014random, lillicrap2016random, liao2016important}.

%To address this issue, different local learning rules have been proposed \cite{movellan1991contrastive, hinton2002training, scellier2017equilibrium, bengio2015towards, lillicrap2020backpropagation}. 

%A second reason that backprop is generally regarded as being biologically implausible is that it relies on symmetric feedback connections for transmitting error signals. However, such symmetric connections have not been widely observed in biological systems. This is also one of the limitations for the learning rules mentioned above, except \cite{bengio2015towards}, which proposed learning the feedback weights instead of assuming symmetric feedback connections.

Alternatively, REINFORCE \cite{williams1992simple} could be applied to all units in the network to train an ANN as a more biologically plausible way of learning. It is shown that the learning rule gives an unbiased estimate of the gradient of return \cite{williams1992simple}. Another interpretation of this relates to viewing each unit as a reinforcement learning (RL) agent, with each agent trying to maximize the global reward. Such a \emph{team of agents} is also known as \emph{coagent network} \cite{thomas2011policy}. However, coagent networks can only solve simple tasks due to the high variance associated with the learning rule and thus the low speed of learning. The high variance stems from the lack of structural credit assignment, i.e.\ a single scalar reward is used to evaluate the action of all agents in the network.

To address this high variance associated with REINFORCE, we propose a novel algorithm that significantly reduces the variance while retaining the local property of the learning rule. We call this newly proposed algorithm \emph{maximum a posteriori (MAP) propagation}. Essentially, MAP propagation replaces the hidden units' values with their MAP estimates conditioned on the action chosen, or equivalently, minimizes the energy function of the network, before applying REINFORCE. We prove that for a network with normally distributed hidden units, by minimizing the energy function of the network, the parameter update given by REINFORCE and backprop (with the reparametrization trick) becomes the same, thus establishing a connection between REINFORCE and backprop. Our experiments show that a team of agents trained with MAP propagation can learn much faster than REINFORCE, such that the team of agents can solve common RL tasks at a similar (or higher) speed compared to an ANN trained by backprop, as well as exhibiting sophisticated exploration that differs from an ANN trained by backprop.

The novel MAP propagation algorithm represents a new class of algorithm to train an ANN that is more biologically plausible than backprop; at the same time, maintaining a comparable learning speed to backprop. Our work also opens the prospect of the broader application of teams of agents, called coagent networks by \cite{thomas2011policy}, in deep RL.

\section{Background and Notation} \label{sec:n}

We consider a Markov Decision Process (MDP) defined by a tuple $(\mathcal{S}, \mathcal{A}, P, R, \gamma, d_0)$, where $\mathcal{S}$ is a finite set of states of an agent's environment (although this work can be extended to the infinite state case), $\mathcal{A}$ is a finite set of actions, $P:\mathcal{S}\times \mathcal{A}\times \mathcal{S} \rightarrow [0,1]$ is a transition function giving the dynamics of the environment, $R: \mathcal{S}\times \mathcal{A} \rightarrow \mathbb{R}$ is a reward function, $\gamma \in [0,1]$ is a discount factor, and $d_0: \mathcal{S} \rightarrow [0,1]$ is an initial state distribution.  Denoting the state, action, and reward signal at time $t$ by  $S_t$, $A_t$, and $R_t$ respectively, $P(s, a, s') = \pr(S_{t+1}=s'|S_t=s, A_t=a)$, $R(s, a) = \ex[R_t|S_t=s, A_t=a]$, and $d_0(s) = \pr(S_{0}=s)$, where $P$ and $d_0$ are valid probability mass functions. An episode is a sequence of states, actions, and rewards, starting from $t=0$ and continuing until reaching the terminal state. For any learning methods, we can measure its performance as it improves with experience over multiple episodes, which makes up a run.

Letting $G_t = \sum_{k=t}^{\infty} \gamma^{k-t} R_k$ denote the infinite-horizon discounted return accrued after acting at time $t$, we are interested in finding, or approximating, a \emph{policy} $\pi:  \mathcal{S}\times \mathcal{A} \rightarrow [0,1]$ such that for any time $t > 0$, selecting actions according to $\pi(s,a)=\pr(A_t=a|S_t=s)$ maximizes the expected return $\ex[G_t|\pi]$. The value function for policy $\pi$ is $V^{\pi}$ where for all $s \in  \mathcal{S}$, $V^{\pi}(s) = \ex[G_t|S_t=s, \pi]$, which can be shown to be independent of $t$ for the infinite-horizon case we are considering. 

Here we restrict attention to policies computed by multi-layer networks consisting of $L$ layers of stochastic units. Let $H^l_t \in \mathbb{R}^{n(l)}$ denotes the activation values of the units in layer $l$ at time $t$ and $n(l)$ denotes the number of units in layer $l$. For any $t > 0$, we also let $H^0_t= S_t$, $H^L_t=A_t$, and $H_t=\{H^1_t, H^2_t, ..., H^{L-1}_t\}$. We call any elements in $H_t$ to be a hidden layer and $H^L_t$ to be the output layer. The distribution of $H^l_t$ conditional on $H^{l-1}_t$ is given by $\pi_l: \mathbb{R}^{n(l-1)} \times \mathbb{R}^{n(l)} \rightarrow [0,1]$, such that for any $t > 0$, $\pi_l(h^{l-1}, h^{l}; W^l) =\pr(H^l_t=h^{l}|H^{l-1}_t=h^{l-1}; W^l)$, where $W^l$ is the parameter of layer $l$. We also denote all parameters of the network as $W = \{W^1, W^2, ..., W^L\}$. To sample an action $A_t$ from the network, we iteratively sample $H^l_t \sim  \pi_l(H^{l-1}_t, \cdot ; W^l)$ from $l=1$ to $L$.

We call layer $l$ to be normally distributed if $\pi_l(H^{l-1}_t,  \cdot; W^l) = N(g^l(H^{l-1}_t; W^{l}), \sigma^2_l)$, the normal distribution with mean $g^l(H^{l-1}_t; W^{l})$, where $g^l:\mathbb{R}^{n(l-1)} \rightarrow \mathbb{R}^{n(l)}$ is a differentiable function, and a fixed standard deviation $\sigma_l$. A common choice of $g$ is a linear transformation followed by an activation function; that is, $g(H^{l-1}_t; W^{l}) = f(W^{l}H^{l-1}_t)$ where $f$ is a non-linear activation function such as softplus or rectified linear unit (ReLU) and $W^{l} \in \mathbb{R}^{n(l) \times n(l-1)}$. We also define the \emph{energy function} $E:\mathbb{R}^{n(1)} \times \mathbb{R}^{n(2)} \times ... \times \mathbb{R}^{n(L-1)} \rightarrow [0, \infty)$ to be $E(h; s, a) = -\log \pr(H_t=h|S_t=s,A_t=a)$, which can be shown to be independent of $t$. 

The case we consider here is one in which all the units of the network implement an RL algorithm and share the same reward signal. These networks can therefore be considered to be \emph{teams of agents} (agent here refers to an \emph{RL agent} \cite{sutton2018reinforcement}), which have also been called \emph{coagent networks} \cite{thomas2011policy}. 

We denote $\nabla_xf$ as the gradient of $f$ w.r.t. $x$, $A^T$ as the transpose of matrix $A$, and $\nabla_A f(\pr(A))$ as the shorthand for $\nabla_a f(\pr(A=a))$. For a random variable $X$ with a distribution that depends on parameter $W$ and a random variable $Y$, we call $h(Z; W, Y)$ to be the \emph{re-parameterization of $X$ by $Z$ conditioned on $Y$} if $h(Z; W, Y)$ and $X$ have the same conditional distribution; that is, $\pr(h(Z; W, Y)=x|Y=y) = \pr(X=x|Y=y; W)$ for all values of $x$, $y$ and $W$, where $Z$ is an independent random variable with a distribution that does not depend on parameter $W$ and $h$ is an invertible and differentiable function. In case $X$ has a multi-layer structure, we denote $h^l(Z; W, Y)$ to be the  $l$\textsuperscript{th} layer in $h(Z; W, Y)$ and $h^{-1}$ as the inverse of $h$. In general, we use the superscript $l$ to denote the $l$\textsuperscript{th} layer in a variable if the variable has a multi-layer structure.  Also, for all distributions discussed in this paper, the probability mass function is replaced by probability density function if the random variable is continuous.

\section{Algorithm}
\subsection{MAP Propagation}

MAP propagation is based on REINFORCE applied to each hidden unit with the same global reinforcement signal. To reduce the variance associated with the learning rule, we note that this variance can be reduced by using the expected parameter update conditioned on the state and the selected action instead. However, this expected parameter update is generally intractable to compute analytically. Therefore, we propose to use the MAP estimate to approximate the expected parameter update. This makes the resulting learning rule biased but reduces the variance significantly. The details of MAP propagation are as below.

The gradient of return with respect to $W^l$ (where $l \in \{1, 2, ..., L\}$ in all discussion below unless stated otherwise) can be estimated by REINFORCE, also known as likelihood ratio estimator:
\begin{align}
	&\nabla_{W^l} \ex[G_t] = \sum_{k=t}^\infty \gamma^{(k-t)} \ex[G_k\nabla_{W^l} \log \pr (A_k|S_k)]. \label{eq:0} 
\end{align}
We can show that the terms in the summation of (\ref{eq:0}) also equals $\ex[G_k\nabla_{W^l}\log \pi_l(H^{l-1}_k, H^l_k; W^l)]$, which is the REINFORCE learning rule applied to each hidden unit with the same global reinforcement signal $G_k$:
\begin{theorem}\label{thm:1}
	Let the policy be a multi-layer network of stochastic units as defined in Section \ref{sec:n}. For any $t >0$ and $l \in \{1, 2, ..., L\}$,
	\begin{align} \ex[G_t\nabla_{W^l}\log \pr(A_t|S_t;W)] =  \ex[G_t\nabla_{W^l}\log  \pi_l(H^{l-1}_t, H^l_t; W^l) ].
	\end{align}
\end{theorem}
The proof is in Appendix B.1. Note that this theorem is also proved in \cite{williams1992simple}. This shows that we can apply REINFORCE to each unit of the network, and the learning rule still gives an unbiased estimate of the gradient of the return. Therefore, denoting $\alpha$ as the step size, we can update parameters by the following stochastic gradient ascent rule:
\begin{equation}
	W^l \leftarrow W^l + \alpha G_t\nabla_{W^l}\log \pi_l(H^{l-1}_t, H^l_t; W^l). \label{eq:1}
\end{equation}
However, this learning rule suffers from high variance since a single reward, which results from the stochastic noise of all units, is used to evaluate actions of all units, making the learning rule scales poorly with the number of units in the network. To reduce the variance, we notice that we can replace $\nabla_{W^l} \log \pi_l(H^{l-1}_t, H^l_t; W^l)$ in learning rule (\ref{eq:1}) by $\ex[\nabla_{W^l}\log \pi_l(H^{l-1}_t, H^l_t; W^l)|S_t, A_t]$, noting that (see Appendix B.1 for the details; at the R.H.S., the outer expectation is taken over $S_t$, $A_t$ and $G_t$, while the inner expectaion is taken over $H^{l-1}_t$ and $H^l_t$):
\begin{equation}
	\ex[G_t\nabla_{W^l}\log  \pi_l(H^{l-1}_t, H^l_t; W^l)] = \ex[G_t \ex[\nabla_{W^l}\log  \pi_l(H^{l-1}_t, H^l_t; W^l)|S_t, A_t]]. \label{eq:3}
\end{equation}
This can reduce variance since the variance associated with the stochastic noise of hidden units is removed in the learning rule (see Appendix B.6 for the proof). The learning rule now becomes:
\begin{equation}
	W^l \leftarrow W^l + \alpha G_t\ex[\nabla_{W^l}\log  \pi_l(H^{l-1}_t, H^l_t; W^l)|S_t, A_t]. \label{eq:1b}
\end{equation}
Since (\ref{eq:1}) is following gradient of return in expectation, and the expected update value of (\ref{eq:1}) and (\ref{eq:1b}) is the same, we conclude that (\ref{eq:1b}) is also a valid learning rule as it follows the gradient of return in expectation. However, we note that $\ex[\nabla_{W^l}\log \pi_l(H^{l-1}_t, H^l_t; W^l)| S_t, A_t]$ in (\ref{eq:1b}) is generally intractable to compute analytically. Instead, we propose to use \emph{maximum a posteriori} (MAP) estimate to approximate this term:\footnote{We let $\hat{H}_t^0=H^0_t=S_t$, $\hat{H}_t^l$ denotes the $l$\textsuperscript{th} layer in $\hat{H}_t$ for $l \in \{1, 2, ..., L-1\}$, and $\hat{H}_t^L=H^L_t=A_t$.}
\begin{equation}
	\ex[\nabla_{W^{l}}\log  \pi_l(H^{l-1}_t, H^l_t; W^l)|S_t, A_t] \approx \nabla_{W^{l}}\log \pi_l(\hat{H}^{l-1}_t, \hat{H}^l_t; W^l), \label{eq:4}
\end{equation} 
where $\hat{H}_t = \mathop{\mathrm{argmax}}_{h_t}   \pr(H_t=h_t|S_t, A_t)$. There are many methods to approximate $\hat{H}_t$, such as hill-climbing methods. In case of hidden units being continuous, we can approximate $\hat{H}_t$ by running gradient ascent on $ \log \pr(H_{t}|S_t, A_t)$ as a function of $H_t$ for fixed $S_t$ and $A_t$, such that $H_t$ approaches $\hat{H_t}$. $H_t$ can be initialized as the value sampled from the network when sampling action $A_t$. To be specific, before applying learning rule (\ref{eq:1}), we first run gradient ascent on $H_t$ for $N$ steps:
\begin{equation}
	H_{t} \leftarrow H_{t} + \alpha \nabla_{H_{t}} \log \pr(H_{t}|S_t, A_t). \label{eq:5}
\end{equation}
For $l \in \{1, 2, ..., L-1\}$, this is equivalent to (see Appendix B.4 for the details):
\begin{align}
	H^l_{t} \leftarrow H^l_{t} + \alpha(\nabla_{H^l_t} \log \pi_{l+1}(H^{l}_t, H^{l+1}_t; W^{l+1}) + \nabla_{H^l_{t}} \log \pi_l(H^{l-1}_t, H^l_t; W^l)).\label{eq:6} 
\end{align}
The update rule is maximizing the probability of the value of a hidden unit given the value of units one layer below and above by updating the value of that hidden unit. Using the definition of \emph{energy function} in Section \ref{sec:n}, then the update rule of hidden units can be seen as minimizing the energy function $E(H_t; S_t,A_t)$ \cite{lecun2006tutorial}.

%In the case of the values of hidden units being normally distributed (that is, $\pi_l(H^{l-1}_t,  \cdot; W^l) = N(g(W^{l}H^{l-1}_t), \sigma^2)$), (\ref{eq:6}) becomes:
%\begin{align}
%H^l_{t} \leftarrow H^l_{t} + \frac{\alpha}{\sigma^2}(g(W^{l}H^{l-1}_t)-H^l_{t}+ ((H^{l+1}_{t}-g(W^{l+1}H^{l}_t)) \odot g'(W^{l+1}H^{l}_t)) (W^{l+1})^T).\label{eq:6-2}
%\end{align}
%We observe that the feedback weight $(W^{l+1})^T$ has to be symmetric of the feedforward weight. 

After updating $H_t$ for $N$ steps by (\ref{eq:6}), we obtain an estimate of $\hat{H}_t$, denoted as $\tilde{H}_t$, and apply the following learning rule to learn the parameters of network:
\begin{equation}
	W^l \leftarrow W^l + \alpha G_t\nabla_{W^l}\log \pi_l(\tilde{H}^{l-1}_t, \tilde{H}^l_t; W^l). \label{eq:9.5}
\end{equation}
We call the algorithm that uses an estimate of $\hat{H}_t$ in the REINFORCE learning rule as \emph{MAP propagation}. The pseudo-code of MAP propagation with gradient ascent to approximate $\hat{H}_t$
can be found in Algorithm 1 in Appendix A. Note that $N=0$ recovers the special case of pure REINFORCE. 
%Note that since $G_t$ is unknown at time $t$, we have to update the parameters at the end of the episode. 

Similar to \emph{actor-critic networks} \cite{sutton2018reinforcement}, we can also train a critic network to estimate the state-value, $V^{\pi}(S_t)$, so $G_t$ in (\ref{eq:9.5}) can be replaced by TD error $\delta_t = R_t + \gamma V^{\pi}(S_{t+1}) - V^{\pi}(S_{t})$ and the whole algorithm can be implemented online. To better facilitate temporal credit assignment, we can also use eligibility traces to replace the gradient in (\ref{eq:9.5}), using the same idea of actor-critic networks with eligibility trace \cite{sutton2018reinforcement}. The pseudo-code of it can be found in Algorithm 2 in Appendix A. 

A team of agents can also be trained by MAP propagation to estimate the state-value, such that a separate team of agents can fulfill the role of a critic network, and the whole actor-critic network can be trained without backprop. A simple way to achieve this is to convert the estimation of state-value to an RL task but this conversion is inefficient since the information of optimal actions is lost (the agent only knows a scalar reward but not the target output). Appendix C proposes a new learning rule to train a team of agents to estimate the state-value by MAP propagation efficiently based on the information of optimal actions.

Essentially, MAP propagation is equivalent to applying REINFORCE after minimizing the energy function. As there are many studies on the biological plausibility of REINFORCE, we refer readers to chapter 15 of \cite{sutton2018reinforcement} for a review and discussion of the connection between REINFORCE and neuroscience. The main difference between MAP propagation and REINFORCE is the minimization of the energy function given by the update rule (\ref{eq:6}). This update rule is \emph{local} as it only depends on the units one layer above and below based on feedforward and feedback connections. There is much evidence that feedback signals in brains alter neural activity \cite{lillicrap2020backpropagation, roelfsema2018control}, supporting the use of feedback connections in MAP propagation. The update rule can also be performed in parallel for all layers, removing the need for precise coordination between feedforward and feedback connections as in backprop. However, the update rule requires the feedback weight to be symmetric of the feedforward weight and different values to be propagated through feedforward and feedback connections. 

MAP propagation fits well into the recently proposed \emph{NGRAD hypothesis} \cite{lillicrap2020backpropagation}, which hypothesizes that the cortex use differences in activity states to drive learning. The main idea of NGRAD is that ``\textit{higher-level activities can nudge lower-level activities towards values that are more consistent with the higher-level activity}'', which also describes the process of energy minimization in MAP propagation. A detailed discussion of the biological plausibility of MAP propagation and its relationship with the NGRAD hypothesis can be found in Appendix F.

\subsection{Relationship with Backpropagation}

A network of stochastic units cannot be directly trained by backprop. However, assuming that there exist a re-parameterization of $H_t$ by $Z_t$ conditioned on $S_t$, denoted by $h(Z_t; W, S_t)$, then we can update parameters using backprop with the re-parameterization trick \cite{kingma2013auto}; that is, for $l \in \{1, 2, ..., L\}$:
\begin{equation}
	W^l \leftarrow W^l + \alpha G_t\nabla_{W^l}\log \pi_L(h^{L-1}(Z_t; W, S_t), A_t; W^L). \label{eq:10}
\end{equation}
It can be shown that this learning rule follows the gradient of return in expectation (See Appendix B.5 for the proof). Using a similar argument as in MAP propagation, we can reduce the variance associated with the learning rule by minimizing the energy function before applying the learning rule. 

Interestingly, for a network with all hidden layers being normally distributed, when the values of hidden layers are settled to a stationary point of the energy function, the parameter update given by backprop (with the reparametrization trick) in (\ref{eq:10}) is equivalent to the parameter update given by REINFORCE in (\ref{eq:1}):\footnote{Similar to footnote 1, we let $\hat{h}^0=s$, $\hat{h}^l$ denotes the $l$\textsuperscript{th} layer in $\hat{h}$ for $l \in \{1, 2, ..., L-1\}$, and $\hat{h}^L=a$.}

\begin{theorem}\label{thm:1.5} 
	Let the policy be a multi-layer network of stochastic units with all hidden layers normally distributed as defined in Section \ref{sec:n}. There exists a re-parameterization of $H_t$ by ${Z}_t$ conditioned on $S_t$ that is independent of $t$, denoted by $h(Z_t; W, S_t)$, such that for any $l \in \{1, 2, ..., L\}$, $s \in \mathcal{S}$, $\hat{h} \in \mathbb{R}^{n(1)} \times \mathbb{R}^{n(2)} \times ... \times \mathbb{R}^{n(L-1)}$, $\hat{z} \in \mathbb{R}^{n(1)} \times \mathbb{R}^{n(2)} \times ... \times \mathbb{R}^{n(L-1)}$ and $a \in \mathcal{A}$, if $\nabla_h E(\hat{h}; s, a) = 0$ and $\hat{z} = h^{-1}(\hat{h}; W, s)$, then
	\begin{align} 
		\nabla_{W^l}\log \pi_l(\hat{h}^{l-1}, \hat{h}^l; W^l) = \nabla_{W^l}\log \pi_L(h^{L-1}(\hat{z}; W, s), a; W^L).
	\end{align}
\end{theorem}

The proof is in Appendix B.2. In other words, by nudging the value of units in lower layers towards values that are more consistent with the value of units in the final layer, the parameter update given by REINFORCE becomes the same as backprop (with the reparametrization trick). With $N \rightarrow \infty$ and $\alpha$ sufficiently small, under update rule (\ref{eq:5}), $H_t$ will converge to the stationary point of energy function. Therefore, the parameter update given by MAP propagation converges to the parameter update given by backprop (with the reparametrization trick) after minimizing the energy function. 

Despite the close relationship between MAP propagation and backprop, there are key differences between the two algorithms. Compared to backprop, one major limitation of MAP propagation is that it can only be applied to RL tasks. MAP propagation is also more computationally expensive than backprop due to the minimizing of the energy function in every step. However, MAP propagation can be applied to a network of discrete units. Moreover, MAP propagation does not require non-local feedback signals or precise coordination between feedforward and feedback pathways, which makes it more biologically plausible than backprop.

To see that MAP propagation is more computationally expensive than backprop, denote  $L$ as the number of layers in the network and $N$ as the number of steps in the energy minimization (the inner loop). MAP propagation requires $LN$ layer updates during each step of energy minimization, while backprop requires $L$ layer updates to compute the feedback signals (iterating from the top layers). Therefore, MAP propagation takes $N$ times more layer updates than backprop. However, the $LN$ layer updates in MAP propagation can be done in parallel for all layers, so the time complexity for a single step of MAP propagation can be reduced to $\mathcal{O}(N)$ from $\mathcal{O}(LN)$ if the update is done in parallel. For backprop, parallel computation of feedback signals is not possible, so the time complexity for a single step of backprop remains $\mathcal{O}(L)$.

\section{Related Work}

Local learning rules based on MAP estimates of latent variables have been proposed in both unsupervised and supervised learning tasks. For unsupervised learning tasks, \cite{bengio2015towards} proposed training a deep generative model by using MAP estimate to infer the value of latent variables, conditioned on observed variables; in the same work, they also proposed to learn the feedback weights such that the constraint of symmetric weight can be removed. This idea can also possibly be applied to our algorithm. For supervised learning tasks,  \cite{whittington2017approximation} proposed training a deep network with local learning rules based on MAP inference and clipping the output value of the network to the target value. In contrast to these works, MAP propagation applies to RL tasks and does not require clipping any units' values.

%\cite{bengio2015early} explained the relationship of the inference of latent variables in energy-based models with backprop.

Besides algorithms based on MAP estimates, many biologically plausible alternatives to backprop have been proposed. \cite{pozzi2020attention, roelfsema2005attention, rombouts2015attention} introduced biologically plausible learning rules based on reward prediction errors and attentional feedback; but these learning rules mostly require a non-local feedback signal. Moreover, \cite{movellan1991contrastive, hinton2002training, scellier2017equilibrium} introduced local learning rules based on contrastive divergence or nudging the values of output units towards the target value. See \cite{lillicrap2020backpropagation} for a comprehensive review of algorithms that approximate backprop with local learning rules based on the differences in units' values. Contrary to \cite{movellan1991contrastive, hinton2002training, scellier2017equilibrium}, MAP propagation requires neither the temporal difference in units' value nor multiple phases of learning. %Also, MAP propagation applies to RL tasks and thus does not assume that the correct output value is known.

Another perspective of training a multi-level network of stochastic units relates to viewing each unit as an RL agent, forming a hierarchy of agents. In hierarchical RL, \cite{levy2017learning} proposed learning a multi-level hierarchy with hindsight actions, which is similar to our idea of replacing the values of hidden layers with the MAP estimates. The special case of a team of agents forming a network to solve a task in a cooperative way was first proposed by \cite{tsetlin1973automaton, barto1985learning}, and a comprehensive review can be found in chapter 15.10 of \cite{sutton2018reinforcement}. Such a team of agents is recently called coagent networks \cite{thomas2011policy} and \cite{thomas2011policy, kostasasynchronous, thomas2011conjugate} introduced theories relating to training coagent networks. However, 
coagent networks learn much slower than an ANN trained by backprop due to the high variance associated with the learning rule. To reduce the variance, \cite{thomas2011policy} proposed to disable exploration of units randomly, but the learning speed is still not comparable to backprop.

In addition, there is a large amount of literature on methods for training a network of stochastic units. A review can be found in \cite{weber2019credit}, which includes the re-parametrization  trick \cite{kingma2014efficient} and REINFORCE \cite{williams1992simple}. They introduced methods to reduce the variance of the estimate, such as baseline and critic. These ideas are orthogonal to the use of MAP estimate to reduce the variance associated with REINFORCE. 

\section{Experiments} 
To test the algorithm, we first consider a single-time-step MDP that is similar to the multiplexer task \citep{barto1985learning}. This is to test the performance of the algorithm as an actor. Then we consider a scalar regression task to test the performance of the algorithm as a critic. Finally, we consider some standard RL tasks to test the performance of the algorithm as both an actor and a critic.

In the below tasks, all the teams of agents (or coagent network) have the same architecture: a two-hidden-layer network, with the first hidden layer having 64 units, the second hidden layer having 32 units, and the output layer having one unit. All hidden layers are normally distributed with $\pi_l(H^{l-1}_t,  \cdot; W^l) = N(f(W^{l}H^{l-1}_t), \sigma^2_l)$ for $l=1, 2$, and $f(x)=\log(1+\exp(x))$, the softplus function. For the network in multiplexer task and the actor network with discrete output, the output unit's distribution is given by the softmax function on the previous layer, i.e.\ $\pi_L(H^{L-1}, a;W^L) = \text{softmax}_a(T  W^L H^{L-1})$, where $T>0$ is a scalar hyperparameter representing the temperature. For the network in the scalar regression task, the critic network and the actor network with continuous output, the output unit's distribution is normally distributed with mean given by a linear transformation of the previous layer's value and a fixed variance, i.e.\ $\pi_L(H^{L-1}, \cdot;W^L) =  N( W^L H^{L-1}, \sigma^2_L)$. We used $N=20$ in MAP propagation. Other hyperparameters and details of experiments can be found in Appendix D.

\begin{figure}%
	\centering
	\subfloat[Multiplexer]{{\includegraphics[width=.5\textwidth]{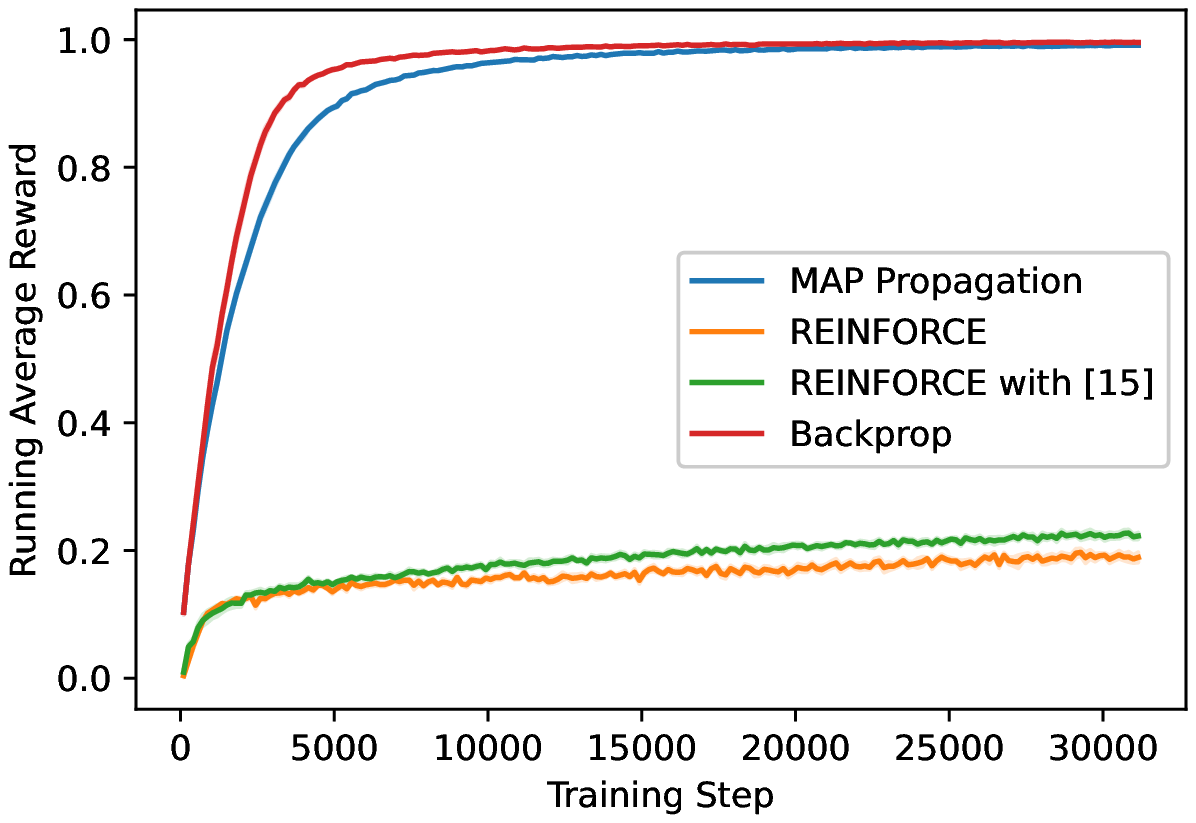} }}%     
	\subfloat[Scalar Regression]{{\includegraphics[width=.5\textwidth]{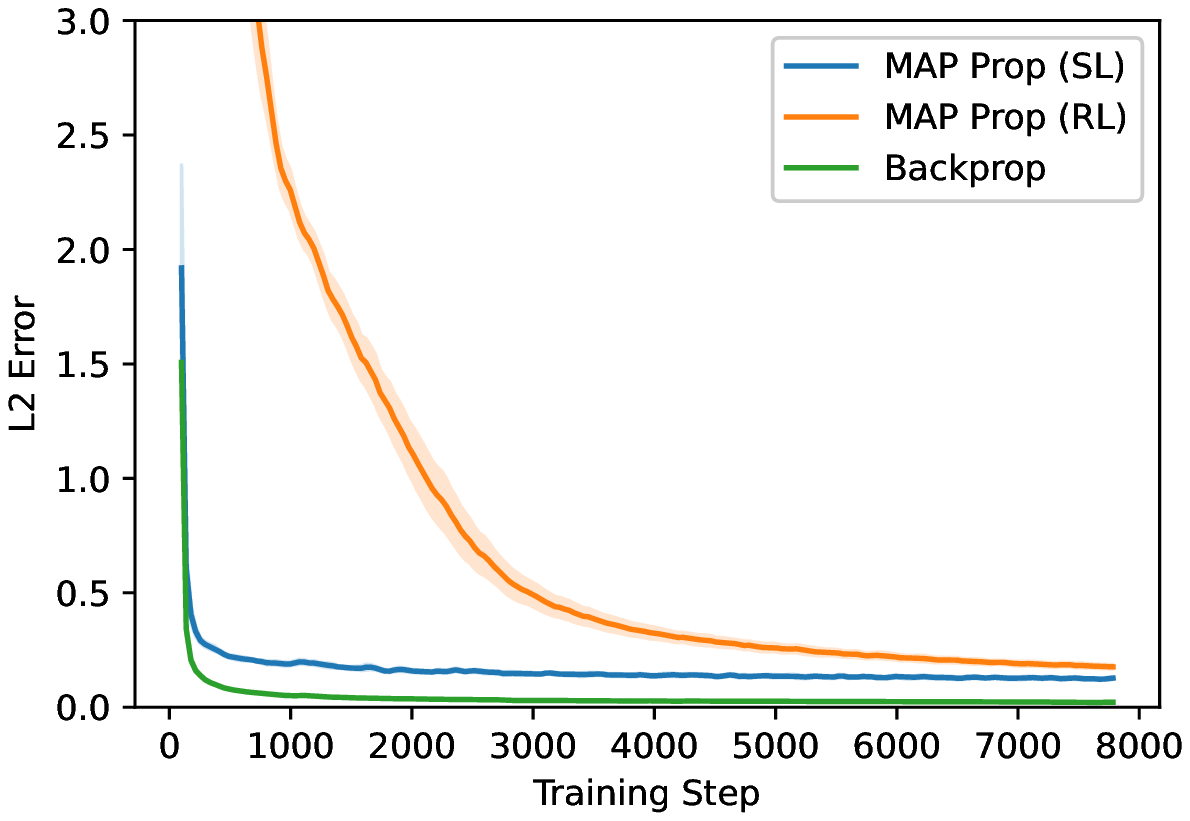} }}%
	\caption{Running average rewards over last 10 episodes in multiplexer task and scalar regression task. Results are averaged over 10 independent runs, and shaded area represents standard deviation over the runs.}%
	\label{fig:1}%
\end{figure}

\begin{figure}%
	\begin{minipage}{.5\linewidth}
		\centering
		\subfloat[Acrobot]{{\includegraphics[width=1\textwidth]{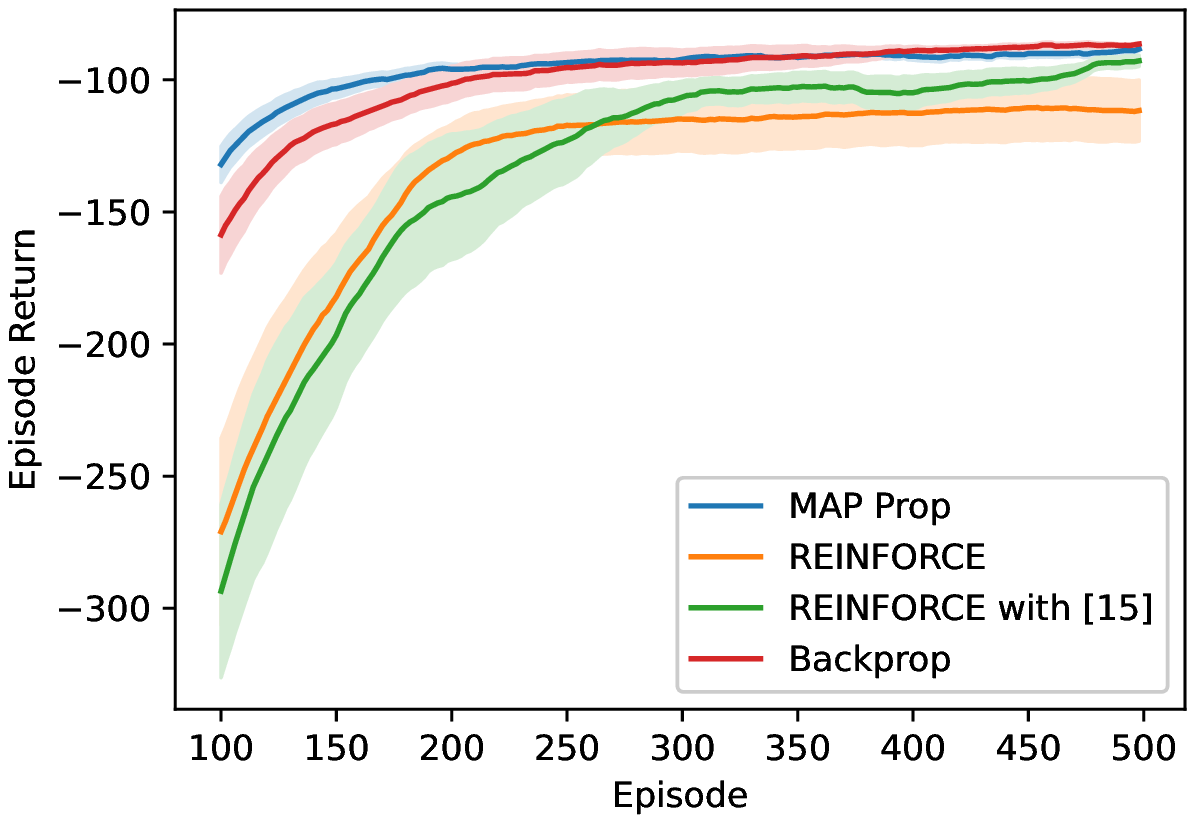} }}%     
	\end{minipage}%
	\begin{minipage}{.5\linewidth}    
		\centering	
		\subfloat[CartPole]{{\includegraphics[width=1\textwidth]{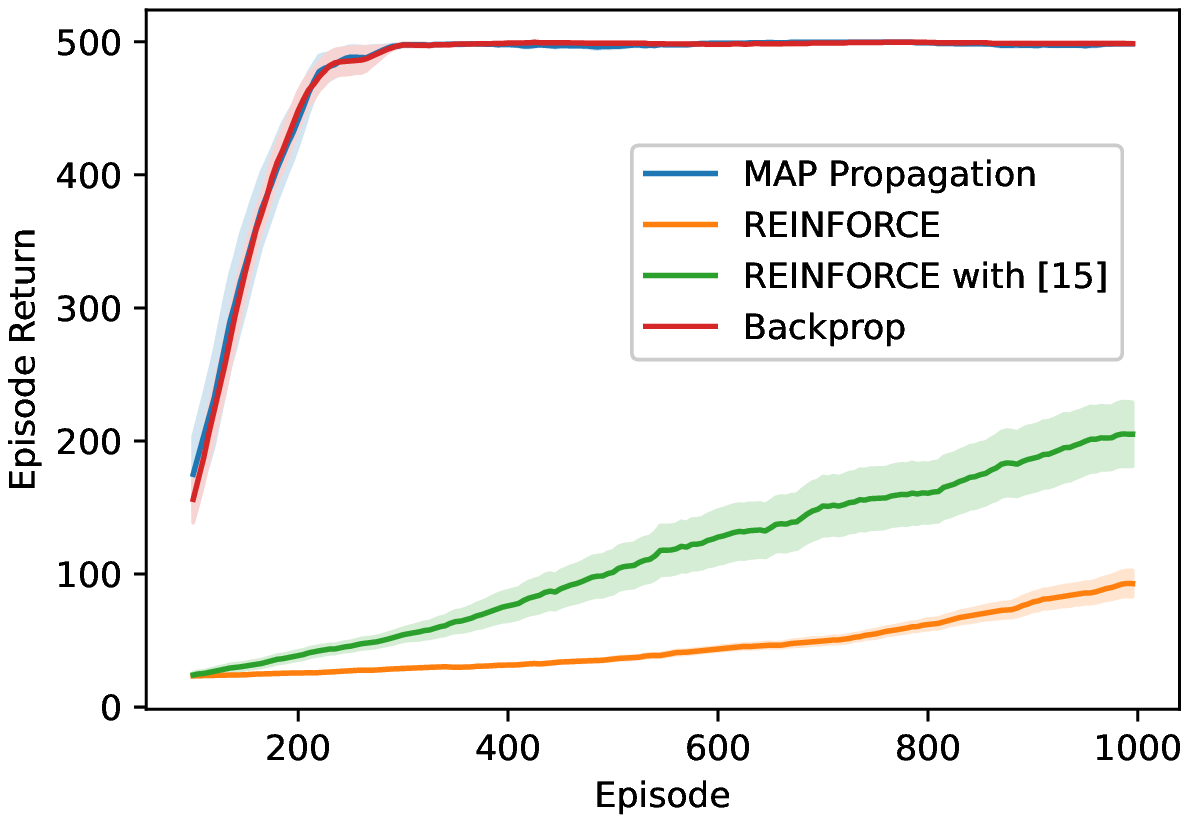}}}%
	\end{minipage}\par\medskip
	\begin{minipage}{.5\linewidth}    
		\centering
		\subfloat[LunarLander]{{\includegraphics[width=1\textwidth]{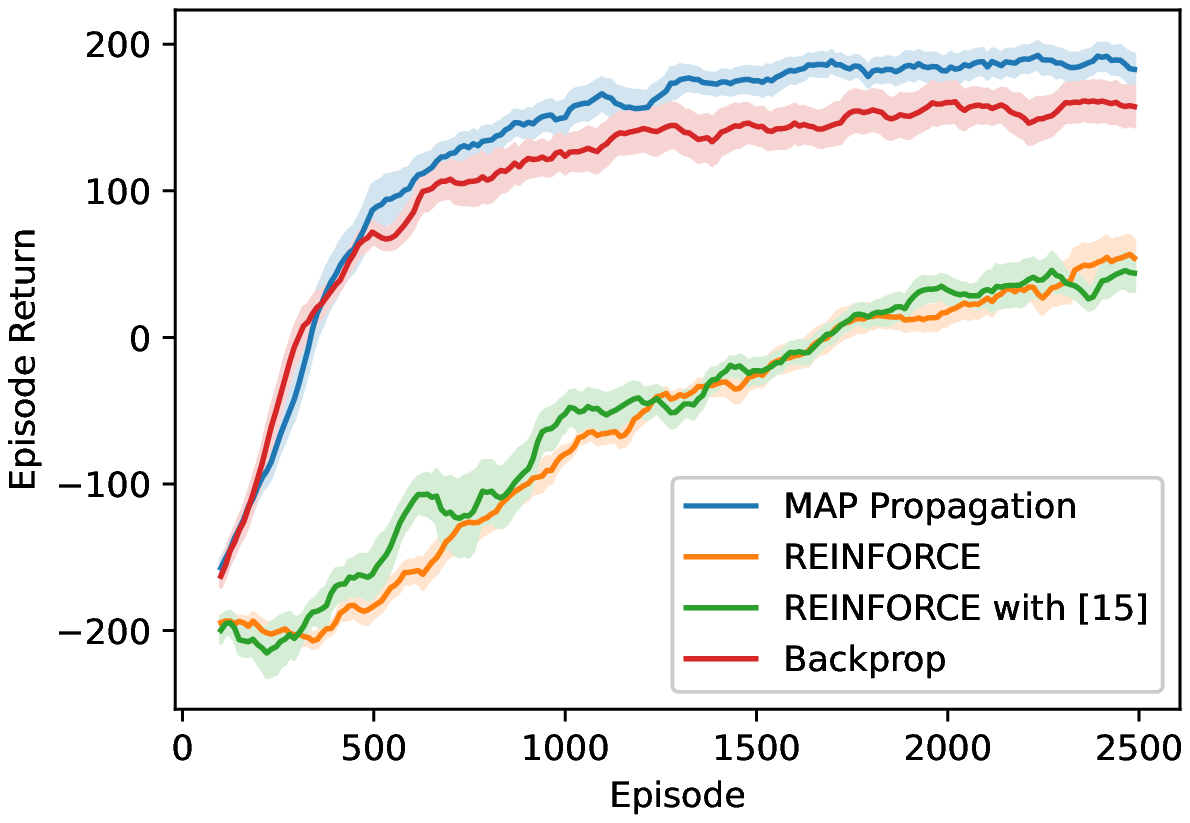} }}%     
	\end{minipage}%
	\begin{minipage}{.5\linewidth}    
		\centering	
		\subfloat[MountainCar]{{\includegraphics[width=1\textwidth]{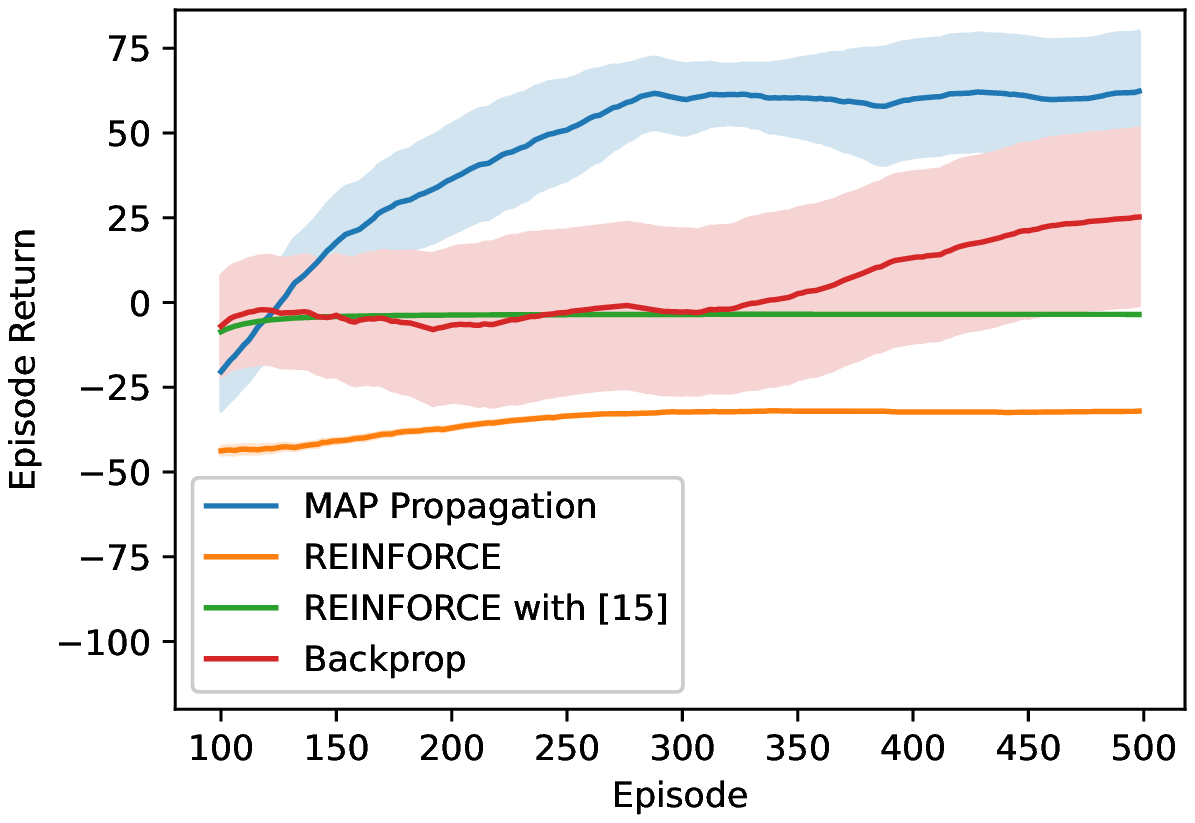}}}%
	\end{minipage}%    
	\caption{Running average returns over the last 100 episodes in Acrobot, CartPole,  LunarLander and MountainCar. Results are averaged over 10 independent runs, and shaded area represents standard deviation over the runs.}%
	\label{fig:2}%
\end{figure}

\begin{table}
	\caption{Average return over all episodes.}
	\label{table:1}
	\centering
	\begin{tabular}{lcccccccccccccc}
		\toprule
		& \multicolumn{2}{c}{Acrobot} & \multicolumn{2}{c}{CartPole} & \multicolumn{2}{c}{LunarLander} & \multicolumn{2}{c}{MountainCar}\\
		\cmidrule(r){2-3}     \cmidrule(r){4-5}  \cmidrule(r){6-7}  \cmidrule(r){8-9}  
		& Mean & Std. & Mean & Std. & Mean & Std. & Mean & Std. \\
		\midrule
		MAP Propagation & \textbf{-100.29} & 5.40 & \textbf{459.70} & 13.89 & \textbf{127.88} & 24.57 & \textbf{39.45} & 30.48\\ 
		REINFORCE & -148.42 & 47.65 & 47.29 & 8.22 & -62.05 & 16.16 & -35.52 & 0.65 \\ 
		REINFORCE with \cite{thomas2011policy} & -149.11 & 33.50 & 112.58 & 42.54 & -54.61 & 23.47 & -4.65 & 0.21 \\ 
		Backprop & -106.29 & 15.00 & 458.96 & 9.44 & 104.92 & 31.98 & 4.30 & 59.28 \\ 
		\bottomrule
	\end{tabular}
\end{table}

\subsection{Multiplexer Task}
We consider a single-time-step MDP that is similar to the $k$-bit multiplexer task. In our single-time-step MDP, the state is sampled from all possible values of a binary vector of size $k + 2^k$ with equal probability. The action set is $\{-1, 1\}$, and we give a reward of 1 if the action of the agent is the desired action and -1 otherwise. The desired action is given by the output of a multiplexer, with the input of the multiplexer being the state. We consider $k=5$ here, so the dimension of the state space is 37.

We used Algorithm 1 for training a team of agents by MAP propagation. We consider three baselines: 1. REINFORCE - A team of agents trained entirely by REINFORCE; 2. REINFORCE with \cite{thomas2011policy} - A team of agents trained entirely by REINFORCE but with the variance reduction method from \cite{thomas2011policy}; 3. Backprop - An ANN with a similar architecture where the output unit is trained by REINFORCE and hidden units are trained by backprop.

The results are shown in Fig \ref{fig:1}. We observe that MAP propagation performs much better than the two REINFORCE baselines. The result suggests that MAP propagation can improve the learning speed of REINFORCE significantly, such that its learning speed is comparable to backprop.

\subsection{Scalar Regression Task}
In the following, we consider a scalar regression task. The dimension of input is 8 and follows the standard normal distribution. The target output (a real scalar) is computed by a one-hidden-layer ANN with weights chosen randomly. The goal of the task is to predict the target output given the input.

For MAP propagation, we used Algorithm 1 and tested two variants. In the first variant that is labeled as `MAP Prop (RL)', we treated the task as a single-time-step MDP with the negative L2 loss as the reward and trained the network using Algorithm 1. In the second variant that is labeled as `MAP Prop (SL)', we replaced the learning rule in Algorithm 1 with the learning rule proposed in Appendix C, which incorporates the value of target output. For the baseline, we trained an ANN with a similar architecture by gradient descent on the L2 loss.

The results are shown in Fig \ref{fig:1}. We observe that if we directly use the negative L2 loss as the reward, then the learning speed of MAP propagation is significantly lower than backprop since the information of target output is not incorporated. On the other hand, if we use the learning rule in Appendix C to incorporate the information of target output, then the learning speed of MAP propagation is comparable to backprop. However, the asymptotic performance of MAP propagation is slightly worse than backprop, which is due to the stochastic property of teams of agents. Nonetheless, this may not be a problem when applying MAP propagation to train a critic network, since the value function to be estimated is also constantly changing with the policy function. 

\subsection{Reinforcement Learning Task}
In the following, we consider four standard RL tasks: Acrobot, CartPole, LunarLander, and continuous MountainCar in OpenAI's Gym. For MAP propagation, we use the actor-critic network with eligibility traces given by Algorithm 2, and the critic network is trained by the learning rule proposed in Appendix C. We consider three baselines. For the first and second baseline, the actor network is a team of agents trained entirely by REINFORCE. However, since REINFORCE cannot train a critic network directly and it is inefficient to convert the state-value estimation task to an RL task, we used an ANN with a similar architecture trained by backprop as the critic network. We also used the variance reduction method from \cite{thomas2011policy} in the second baseline. We used eligibility traces in the training of both the actor and the critic network. For the third baseline, we used actor-critic with eligibility traces (episodic) \cite{sutton2018reinforcement} trained by backprop. Both actor and critic networks are ANNs with an architecture similar to the team of agents. 

The average return over ten independent runs is shown in Fig \ref{fig:2}. Let $\bar{G}$ denote the average return of all episodes. The mean and standard deviation of $\bar{G}$ over the ten runs can be found in Table \ref{table:1}. For all RL tasks, we observe that MAP propagation has a better performance than the baselines in terms of the average return $\bar{G}$. The result demonstrates that a team of agents trained with MAP propagation can learn much faster than a team of agents trained with REINFORCE, such that the team of agents can solve common RL tasks at a similar (or higher) speed compared to an ANN trained by backprop.

Although there are other algorithms besides actor-critic networks that can solve RL tasks more efficiently, the present work aims to compare different training methods for hidden units in an actor-critic network. Teams of agents trained by MAP propagation can also be applied to algorithms besides actor-critic networks, such as variants of actor-critic networks like Proximal Policy Optimization \cite{schulman2017proximal} and action-value based methods like Q-Learning \cite{sutton2018reinforcement}.

We also notice that MAP Prop performs better than backprop on tasks where a high degree of exploration is required. For example, MAP propagation performs slightly worse than backprop on the multiplexer task but much better than backprop on the MountainCar task. This may suggest that teams of agents trained with MAP propagation can have better exploration than an ANN trained with backprop. This is further corroborated by the analysis of agents' behaviors on MountainCar, a task where an agent can easily be stuck in the local optima. For backprop, we found that agents in most of the runs are stuck in early episodes even with the use of entropy regularization \cite{mnih2016asynchronous}. In contrast, a team of agents trained by MAP propagation can reach the goal of the task successfully in all runs. A detailed analysis of this can be found in Appendix E. One possible explanation for the better exploration is that actions of agents in lower layers can be considered as abstract actions, and the exploration of these agents corresponds to exploration beyond the primitive actions.

%In the experiments above, we plotted the running average of returns against the number of the episode. However, it should be noted that the computation required for MAP propagation and backprop in each step is different since the time complexity of each step in MAP propagation is $O(N)$. This makes training of MAP Propagation take a much longer time than backprop. However, one advantage of MAP propagation is that the update rule of (\ref{eq:6}) can be implemented asynchronously, while backprop requires all the upper layers to finish computing the gradient before updating the next layer. 

%Another limitation of MAP propagation is that it is sensitive to a large number of hyperparameters. For example, we found that the variance has to be larger, and the learning rate has to be smaller for units in deeper layers. This is in contrast to the baseline, which only has a few hyperparameters and does not require separate hyperparameters per layer. Further work can be done on investigating the choices of hyperparameters in MAP propagation.

\section{Future Work and Conclusion}
%Much research has been done on the theoretical property of teams of agents, or coagent networks \cite{thomas2011policy, kostasasynchronous, thomas2011conjugate, barto1985learning}, but so far, there are no researches tackling its very low learning speed, a critical problem that makes it not readily scalable to large networks.

The ability to train teams of agents efficiently leads to many possible future directions. First, the local property of the learning rule points to the possibility of implementing MAP propagation asynchronously, such that it can be implemented efficiently with neuromorphic circuits \cite{indiveri2011neuromorphic}. Second, agents in the team can have a different temporal resolution, such that the actions of agents can be extended temporally and become options \cite{sutton1999between}, yielding better exploration and learning behavior.

In conclusion, we propose a new algorithm that reduces the variance associated with REINFORCE and thus significantly increases the learning speed in the training of a team of agents. The proposed algorithm is also more biologically plausible than backprop while maintaining a comparable learning speed to backprop. Our work opens the prospect of the broader application of teams of agents in deep RL. Our experiments also suggest a team of agents trained by MAP propagation can have more sophisticated exploration compared to an ANN trained by backprop.

\section{Acknowledgment}
We would like to thank Andrew G. Barto, who inspired this research and provided valuable insights and comments, as well as Andy K.P. Chan for feedback and discussions.

\bibliographystyle{unsrt}
\bibliography{citation}

	\appendix
	\section{Algorithms} \label{sec:A0}
	
	\begin{algorithm}
		\textbf{Input:} differentiable policy function: $\pi_l(h^{l-1}, h^{l}; W^{l})$  for $l \in \{1, 2, ..., L\}$\;
		\textbf{Algorithm Parameter:} step size $\alpha >0$, $\alpha_{h} > 0$; update step $N \geq 1$\; discount rate $\gamma \in [0,1]$\;
		\textbf{Initialize policy parameter:} $W^{l}$ for {$l \in \{1, 2, ..., L\}$}\;
		
		\SetKwProg{lf}{Loop forever (for each episode):}{}{}
		\SetKwProg{lw}{Loop for each step of the episode, $t=0, 1, ..., T-1$:}{}{}
		\lf{}{  
			Generate an episode $S_0, H_0, A_0, R_1, ..., S_{T-1}, H_{T-1}, A_{T-1}, R_T$ following $\pi_l(\cdot, \cdot; W^{l})$ for $l \in \{1, 2, ..., L\}$\;  
			\lw{}{
				$G \leftarrow \sum_{k=t+1}^T \gamma^{k-t-1}R_k$ \;
				\tcc{MAP Gradient Ascent}             
				
				\For{$n:=1, 2, ..., N$}{       	
					$H^{l}_t \leftarrow H^{l}_t + \alpha_{h} (\nabla_{H^{l}_t} \log \pi_{l}(H^{l-1}_t, H^{l}_t; W^{l}) + \nabla_{H^{l}_t} \log \pi_{l+1}(H^{l}_t, H^{l+1}_t; W^{l+1}))$  for {$l  \in  \{1, 2, ..., L-1\}$}\;     
				}      
				
				\tcc{Apply REINFORCE}       
				$W^{l} \leftarrow W^{l} + \alpha G \nabla_{W^{l}} \log \pi_{l}(H^{l-1}_t, H^{l}_t; W^{l})$ for {$l  \in  \{1, 2, ..., L\}$}\;
				
			}
		}
		
		\caption{MAP Propagation - Monte-Carlo Policy-Gradient Control} \label{alg:1}
	\end{algorithm}
	
	\begin{algorithm}
		\textbf{Input:} differentiable policy function: $\pi_l(h^{l-1}, h^{l}; W^{l})$  for $l \in \{1, 2, ..., L\}$\;
		\textbf{Algorithm Parameter:} step size $\alpha >0$, $\alpha_{h} > 0$; update step $N \geq 1$\; trace decay rate $\lambda \in [0,1]$; discount rate  $\gamma \in [0,1]$\;
		\textbf{Initialize policy parameter:} $W^{l}$  for {$l \in \{1, 2, ..., L\}$}\;
		\SetKwProg{lf}{Loop forever (for each episode):}{}{} 
		\SetKwProg{lw}{Loop while $S$ is not terminal (for each time step):}{}{}
		\lf{}{
			Initialize $S$ (first state of episode) \;
			Initialize zero eligibility trace $\mathbf{z^l}$   for {$l \in \{1, 2, ..., L\}$} \;
			\lw{}{
				$H^{0} \leftarrow S$ \;    
				\tcc{1. Feedforward phase}  	 
				Sample $H^{l}$ from $\pi_{l}(H^{l-1}, \cdot; W^{l})$  for {$l \in \{1, 2, ..., L\}$} \; 	
				$A \leftarrow H^{L}$\;
				
				\tcc{2. REINFORCE phase}      
				\If {\normalfont{\textbf{episode not in first time step}}}{ 
					Receive TD error $\delta$ from the critic network\;
					$W^{l} \leftarrow W^{l} + \alpha \delta \mathbf{z^l}$  for {$l \in \{1, 2, ..., L\}$}{\;	    
					}
				}
				\tcc{3. Minimize energy phase}             
				
				\For{$n:=1, 2, ..., N$}{ 
					$H^{l} \leftarrow H^{l} + \alpha_h (\nabla_{H^{l}} \log \pi_{l}(H^{l-1}, H^{l}; W^{l}) + \nabla_{H^{l}} \log \pi_{l+1}(H^{l}, H^{l+1}; W^{l+1}))$ for {$l \in \{1, 2, ..., L-1\}$} \;   
				}		
				\tcc{4. Trace accumluation phase}     
				$\mathbf{z}^{l}  \leftarrow \gamma \lambda \mathbf{z}^{l} + \nabla_{W^{l}} \log \pi_{l}(H^{l-1}, H^{l}; W^{l})$ for {$l \in \{1, 2, ..., L\}$}\;	
				Take action $A$, observe $S, R$ \;
			}{}
		}{}
		
		\caption{MAP Propagation - Actor Network with Eligibility Trace} \label{alg:2}
	\end{algorithm}

	\section{Proof}
	In the proofs below, we may omit the subscript $t$ whenever it is unnecessary. In addition to the notation in Section \ref{sec:n}, we define $D_xf$ as the Jacobian matrix of $f$ w.r.t. $x$.
	
	\subsection{Proof of Theorem 1} \label{sec:A1}
	\begin{align}
		& \ex[G\nabla_{W^l}\log \pr(A|S; W)] \label{eq:A1}  \\
		=& \ex[\frac{G}{\pr(A|S; W)}\nabla_{W^l}\pr(A|S; W)] \label{eq:A2}\\
		=& \ex[\frac{G}{\pr(A|S; W)}\nabla_{W^l} \sum_{h} \pi_L(h^{L-1}, A; W^L) \pi_{L-1}(h^{L-2}, h^{L-1}; W^{L-1}) 
		...\nonumber \\ &\pi_{2}(h^1, h^2; W^2) \pi_{1}(S, h^1; W^1)] \label{eq:A3}\\
		=& \ex[\frac{G}{\pr(A|S; W)} \sum_{h^{l}, h^{l-1}} \pr(A, H^{l}=h^{l}, H^{l-1}=h^{l-1} | S) \nabla_{W^l} \log \pi_{l}(h^{l-1}, h^l; W^l)] \label{eq:A4} \\
		=& \ex[G \sum_{h^{l-1}, h^l} \pr(H^{l-1}=h^{l-1}, H^l=h^l | S, A)\nabla_{W^l} \log \pi_{l}(h^{l-1}, h^l; W^l)] \label{eq:A5} \\
		=& \ex[G \ex[\nabla_{W^l} \log \pi_{l}(H^{l-1}, H^l; W^l)|S, A]]  \label{eq:A6} \\
		=&\ex[\ex[G|S,A] \ex[\nabla_{W^l} \log \pi_{l}(H^{l-1}, H^l; W^l)|S, A]] \label{eq:A7}\\
		=& \ex[\ex[G \nabla_{W^l} \log \pi_{l}(H^{l-1}, H^l; W^l)|S, A]]  \label{eq:A8}\\
		=& \ex[G \nabla_{W^l} \log \pi_{l}(H^{l-1}, H^l; W^l)]. \label{eq:A9}
	\end{align}
	%(\ref{eq:A2}) to (\ref{eq:A3}) uses the fact that:
	%\begin{align}
	%&\pr(A|S; W) 
	%&= \sum_{h} \pi_L(h^{L-1}, A; W^L) \pi_{L-1}(h^{L-2}, h^{L-1}; W^{L-1}) ... \pi_{2}(h^1, h^2; W^2) \pi_{1}(S, h^1; W^1).
	%\end{align}
	%which marginalizes over all possible value of hidden units $H^1, H^2, ..., H^{L-1}$. \\
	%(\ref{eq:A3}) to (\ref{eq:A4}) uses the fact that:
	%\begin{align}
	%&\nabla_{W^l} \sum_{h} \pi_L(h^{L-1}, A; W^L) \pi_{L-1}(h^{L-2}, h^{L-1}; W^{L-1}) ... \pi_{2}(h^1, h^2; W^2) \pi_{1}(S, h^1; W^1)\\
	%=& \sum_{h^{l+1}, h^l} \frac{\nabla_{W^l} \pi_{l}(h^{l-1}, h^l; W^l)}{\pi_{l}(h^{l-1}, h^l; W^l)}  \sum_{h^1, ..., h^{l-1}, h^{l+2}, ..., h^{L-1}} \pi_L(h^{L-1}, A; W^L) \nonumber\\ &\pi_{L-1}(h^{L-2}, h^{L-1}; W^{L-1}) ... \pi_{2}(h^1, h^2; W^2) \pi_{1}(S, h^1; W^1) \label{eq:B1}\\
	%=&\sum_{h^{l+1}, h^l} \pr(A, H^{l+1}=h^{l+1}, H^l=h^l | S) \nabla_{W^l} \log \pi_{l}(h^{l-1}, h^l; W^l). \label{eq:B2}
	%\end{align}
	%For (\ref{eq:B1}), we first move the term $\pi_{l}(h^{l-1}, h^l; W^l)$ out from inner summation since it is the only term that depends on $W^l$. Then we multiply back $\frac{\pi_{l}(h^{l-1}, h^l; W^l)}{\pi_{l}(h^{l-1}, h^l; W^l)}$ to get (\ref{eq:B1}). (\ref{eq:B2}) uses the fact that the inner summation in (\ref{eq:B1}) is just $\pr(A, H^{l+1}=h^{l+1}, H^l=h^l | S)$.\\
	%(\ref{eq:A4}) to (\ref{eq:A5}) uses the fact that $\frac{ \pr(A, H^{l+1}=h^{l+1}, H^l=h^l | S) }{\pr(A|S)} = \pr(H^{l+1}=h^{l+1}, H^l=h^l | S, A)$.\\
	%(\ref{eq:A5}) to (\ref{eq:A6}) uses the definition of expectation.\\ 
	(\ref{eq:A6}) to (\ref{eq:A7}) uses the fact that, for any random variables $Z$ and $Y$, $\ex[\ex[Z|Y]f(Y)] = \ex[Zf(Y)]$. \\%In our case, $Z = G$, $Y = (S, A)$ and $f(Y) = \ex[\nabla_{W^l}\log \pr(H^l|H^{l-1})|S, A]$. \\
	(\ref{eq:A7}) to (\ref{eq:A8}) uses the fact that $G$ is conditional independent of $H^l, H^{l-1}$ given $S, A$.\\
	(\ref{eq:A8}) to (\ref{eq:A9}) uses the law of total expectation.%: $\ex_{X, Y}[f(X, Y)] = \ex_{Y}[\ex_{X}[f(X, Y)|Y]]$. In our case, $X = (G, H^l, H^{l-1})$, $Y = (S, A)$ and $f(X, Y) = G\nabla_{W^l}\log \pr(H^l|H^{l-1})$. 
	
	Note that (\ref{eq:A6}) to (\ref{eq:A9}) also shows the steps for (\ref{eq:3}).
	
	\setcounter{equation}{20}
	
	\subsection{Proof of Theorem 2} \label{sec:A1.5}
	Using $\nabla_h E(\hat{h};s, a) = 0$, $\hat{h}^{L-1}$ can be expressed as:
	\begin{align} 
		\nabla_h E(\hat{h};s, a) &= 0, \\
		-\nabla_{h^{L-1}} \log \pi(\hat{h}^{L-2}, \hat{h}^{L-1}; W^{L-1}) &= \nabla_{h^{L-1}} \log \pi(\hat{h}^{L-1}, a; W^L),\\
		\frac{1}{(\sigma^{L-1})^2}(\hat{h}^{L-1} - g^{L-1}(\hat{h}^{L-2};W^{L-1})) &= \nabla_{h^{L-1}} \log \pi(\hat{h}^{L-1}, a; W^L),\\
		\hat{h}^{L-1} =  g^{L-1}(\hat{h}^{L-2};W^{L-1}) + &(\sigma^{L-1})^2 \nabla_{h^{L-1}} \log \pi(\hat{h}^{L-1}, a; W^L).
	\end{align}
	And for $l = 1, 2, ..., L-2$, we have: 
	\begin{align} 
		\nabla_h E(\hat{h};s, a) &= 0, \\
		-\nabla_{h^{l}} \log \pi(\hat{h}^{l-1}, \hat{h}^{l}; W^{l}) &= \nabla_{h^{l}} \log \pi(\hat{h}^{l}, \hat{h}^{l+1}; W^{l+1}),\\
		\frac{1}{(\sigma^{l})^2}(\hat{h}^{l} - g^l(\hat{h}^{l-1};W^{l})) &=  \frac{1}{(\sigma^{l+1})^2}D_{h^l} g^{l+1}(\hat{h}^{l};W^{l+1})^T (\hat{h}^{l+1} - g^{l+1}(\hat{h}^{l};W^{l+1})) ,
	\end{align}
	\begin{align}
		\hat{h}^{l} = &\; g^l(\hat{h}^{l-1};W^{l}) + \left( \frac{\sigma^{l}}{\sigma^{l+1}} \right)^2 D_{h^l} g^{l+1}(\hat{h}^{l};W^{l+1})^T (\hat{h}^{l+1} - g^{l+1}(\hat{h}^{l};W^{l+1})) , \\
		\hat{h}^{l} = &\; g^l(\hat{h}^{l-1};W^{l}) + \left( \frac{\sigma^{l}}{\sigma^{l+2}} \right)^2  D_{h^l} g^{l+1}(\hat{h}^{l};W^{l+1})^T D_{h^{l+1}} g^{l+2}(\hat{h}^{l+1};W^{l+2})^T  \nonumber \\  
		&\;  (\hat{h}^{l+2} - g^{l+2}(\hat{h}^{l+1};W^{l+2})), \\
		\hat{h}^{l} = &\; g^l(\hat{h}^{l-1};W^{l}) + (\sigma^{l})^2 D_{h^l} g^{l+1}(\hat{h}^{l};W^{l+1})^T D_{h^{l+1}} g^{l+2}(\hat{h}^{l+1};W^{l+2})^T \nonumber \\
		&\; ... D_{h^{L-2}} g^{L-1}(\hat{h}^{L-2};W^{L-1})^T \nabla_{h^{L-1}}  \log \pi_L(\hat{h}^{L-1}, a; W^L).
	\end{align}
	Substituting back to the REINFORCE update, we have:
	\begin{align}
		&\nabla_{W^l}\log \pi_l(\hat{h}^{l-1}, \hat{h}^l; W^l)\\
		= &\frac{1}{(\sigma^{l})^2} D_{W^l} g^{l}(\hat{h}^{l-1};W^{l})^T (\hat{h}^{l} - g^{l}(\hat{h}^{l-1};W^{l})) \\
		= & D_{W^l} g^{l}(\hat{h}^{l-1};W^{l})^T D_{h^l} g^{l+1}(\hat{h}^{l};W^{l+1})^T D_{h^{l+1}} g^{l+2}(\hat{h}^{l+1};W^{l+2})^T ...  \nonumber \\
		&\; D_{h^{L-2}} g^{L-1}(\hat{h}^{L-2};W^{L-1})^T \nabla_{h^{L-1}}  \log \pi_L(\hat{h}^{L-1}, a; W^L). \label{eq:KN1}
	\end{align}
	We will show that the R.H.S. also equals (\ref{eq:KN1}). Consider the re-parameterization of $H^l $ by $Z^l$ conditioned on $H^{l-1}$ using $g^l(H^{l-1};W^l) + \sigma^l Z^l$ and $Z^l$ are independent standard Gaussian noises for $l \in \{1, 2, ..., L-1\}$. Then by re-parameterizing all hidden layers, we can find $h(Z; W, S)$ that is the re-parameterization of $H$ by $Z:=\{Z^1, Z^2, ..., Z^{L-1}\}$ conditioned on $S$. To be concrete, $h^l(Z; W, S) = g^{l}(h^{l-1}(Z; W, S);W^{l}) + \sigma^{l} Z^{l}$ for $l \in \{1, 2, ..., L-1\}$, and $h^0(Z; W, S) := S$.
	
	Then, for $l \in \{1, 2, ..., L-2\}$, we have:
	\begin{align}
		& \nabla_{W^l}\log \pi_L(h^{L-1}(z; W, s), a; W^L) \label{eq:KN2}  \\
		= & D_{W^l}(h^{L-1}(z; W, s))^T \nabla_{h^{L-1}} \log \pi_L(h^{L-1}(z; W, s), a; W^L)  \\
		= & D_{W^l}(g^{L-1}(h^{L-2}(z; W, s);W^{L-1}) + \sigma^{L-1} z^{L-1})^T \nabla_{h^{L-1}} \log \pi_L(h^{L-1}(z; W, s), A; W^L) \\
		= &D_{W^l} (h^{L-2}(z; W, s))^T D_{h^{L-2}}g^{L-1}(h^{L-2}(z; W, s);W^{L-1})^T \nonumber \\ &\nabla_{h^{L-1}} \log \pi_L(h^{L-1}(z; W, s), a; W^L)   \\
		= &D_{W^l} g^{l}(h^{l-1}(z; W, s);W^{l})^T D_{h^l} g^{l+1}(h^{l}(z; W, s);W^{l+1})^T D_{h^{l+1}} g^{l+2}(h^{l+1}(z; W, s);W^{l+2})^T ... \nonumber \\  
		& D_{h^{L-2}} g^{L-1}(h^{L-2}(z; W, s);W^{L-1})^T \nabla_{h^{L-1}} \log \pi_L(h^{L-1}(z; W, s), a; W^L). \label{eq:KN3} 
	\end{align}
	If we evaluate (\ref{eq:KN2}) at $z = \hat{z}$, then (\ref{eq:KN3}) becomes (\ref{eq:KN1}), which completes the proof for $l \in \{1, 2, ..., L-1\}$. The proof for $l=L$ is similar and is omitted here.
	
	\subsection{Proof of Theorem \ref{thm:3}} \label{sec:A2}
	Similar to the proof of Theorem \ref{thm:1.5}, for $l \in \{1, 2, ..., L-1\}$, the L.H.S. can be expressed as:
	\begin{align} 
		& \frac{A^*(s)-\hat{\mu}^L}{a-\hat{\mu}^L} \nabla_{W^l}\log \pi_l(\hat{h}^{l-1}, \hat{h}^l; W^l) \\
		= & \frac{A^*(s)-\hat{\mu}^L}{a-\hat{\mu}^L} D_{W^l} g^{l}(\hat{h}^{l-1};W^{l})^T D_{h^l} g^{l+1}(\hat{h}^{l};W^{l+1})^T D_{h^{l+1}} g^{l+2}(\hat{h}^{l+1};W^{l+2})^T ...  \nonumber \\
		&\; D_{h^{L-2}} g^{L-1}(\hat{h}^{L-2};W^{L-1})^T \nabla_{h^{L-1}}  \log \pi_L(\hat{h}^{L-1}, a; W^L)\\
		= &\frac{A^*(s)-\hat{\mu}^L}{(\sigma^L)^2} D_{W^l} g^{l}(\hat{h}^{l-1};W^{l})^T D_{h^l} g^{l+1}(\hat{h}^{l};W^{l+1})^T D_{h^{l+1}} g^{l+2}(\hat{h}^{l+1};W^{l+2})^T ...  \nonumber \\
		&\; D_{h^{L-2}} g^{L-1}(\hat{h}^{L-2};W^{L-1})^T \nabla_{h^{L-1}}  g^{L}(\hat{h}^{L-1};W^{L}). \label{eq:P0}
	\end{align}
	We then prove that the R.H.S. also equals (\ref{eq:P0}). Consider the same re-parameterization of $H$ by $Z:=\{Z^1, Z^2, ..., Z^{L-1}\}$ conditioned on $S$, denoted by $h(Z; W, S)$, as in the proof of Theorem \ref{thm:1.5}. Then, for $l \in \{1, 2, ..., L-1\}$, 
	\begin{align}
		& \nabla_{W^l} -\left(A^*(s) -  g^L(h^{L-1}(z; W, s); W^{L}) \right)^2 \label{eq:P1} \\
		= & 2(A^*(s) - g^L(h^{L-1}(z; W, s); W^{L})) \nabla_{W^l}  g^L(h^{L-1}(z; W, s); W^{L})\\
		= & 2(A^*(s) - g^L(h^{L-1}(z; W, S); W^{L}))  D_{W^l} (h^{L-1}(z; W, s))^T \nabla_{h^{L-1}}  g^L(h^{L-1}(z; W, s); W^{L}) \\
		= & 2(A^*(s) - g^L(h^{L-1}(z; W, S); W^{L})) D_{W^l} g^{l}(h^{l-1}(z; W, s);W^{l})^T D_{h^l} g^{l+1}(h^{l}(z; W, s);W^{l+1})^T \nonumber \\ 
		& D_{h^{l+1}} g^{l+2}(h^{l+1}(z; W, s);W^{l+2})^T  ...\;   \nabla_{h^{L-1}} g^{L}({h}^{L-1}(z; W, s);W^{L}). \label{eq:P2}  
	\end{align}
	If we evaluate (\ref{eq:P1}) at $z = \hat{z}$, then (\ref{eq:P2}) becomes proportional to (\ref{eq:P0}) with a ratio $2(\sigma^L)^2$, which completes the proof for $l \in \{1, 2, ..., L-1\}$. The proof for $l=L$ is similar and is omitted here.
	
	\subsection{Details of (\ref{eq:5}) to (\ref{eq:6})}   \label{sec:A2.5}
	
	\begin{align}
		& \nabla_{H^l_{t}} \log \pr(H_{t}|S_t, A_t)  \label{eq:I0}\\
		=& \nabla_{H^l_{t}} \left(\log \pr(H_{t}, A_t|S_t) - \log \pr(A_t|S_t) \right) \label{eq:I1}\\
		=& \nabla_{H^l_{t}} \log \pr(H_{t}, A_t|S_t)  \label{eq:I2}\\
		=& \nabla_{H^l_{t}}  \log  \left(\prod_{i=0}^{L-1} \pi_i(H^{i}_{t}, H^{i+1}_{t}; W^{i+1}) \right)  \label{eq:I3}\\
		=& \nabla_{H^l_t} \log \pi_{l+1}(H^{l}_t, H^{l+1}_t; W^{l+1}) + \nabla_{H^l_{t}} \log \pi_l(H^{l-1}_t, H^l_t; W^l). \label{eq:I5}
	\end{align}
	\subsection{(\ref{eq:10}) Follows the Gradient of Return in Expectation}   \label{sec:A2.6}
	We will show that the learning rule given by (\ref{eq:10}) follows the gradient of return in expectation. For $l \in \{1, 2, ..., L-1\}$:
	\begin{align}
		& \nabla_{W^{l}}\ex[G] \label{eq:M0} \\
		= & \ex[G\nabla_{W^l}\log \pr(A|S; W)] \label{eq:M1}  \\
		=& \ex[\frac{G}{\pr(A|S; W)}\nabla_{W^l}\pr(A|S; W)] \\
		=& \ex[\frac{G}{\pr(A|S; W)}\nabla_{W^l}\sum_z\pr(A|Z=z, S; W) \pr(Z=z| S)] \\
		=& \ex[\frac{G}{\pr(A|S; W)}\sum_z \pr(A|Z=z, S; W)\pr(Z=z| S) \nabla_{W^l} \log\pr(A|Z=z, S; W)]\\
		=& \ex[G\sum_z \pr(Z=z|S,A; W)\nabla_{W^l} \log\pr(A|Z=z, S; W)]\\
		=& \ex[G \ex[\nabla_{W^l} \log\pr(A|Z, S; W)|S, A]]\\
		=& \ex[G \nabla_{W^l}\log\pr(A|Z, S; W)]\\
		=& \ex[G \nabla_{W^l}\log \pi_L(h^{L-1}(Z; W, S), A; W^L)].
	\end{align}
	(\ref{eq:M0}) to (\ref{eq:M1}) uses REINFORCE and other steps are similar to those in the proof of Theorem \ref{thm:1}. 
	
	\subsection{Variance Reduction of (\ref{eq:3})}   \label{sec:A2.6}
	
	We will show that for $l \in \{1, 2, ..., L\}$:
	\begin{equation}
		\var[G \ex[\nabla_{W^l}\log  \pi_l(H^{l-1}, H^l; W^l)|S, A]] \leq \var[G\nabla_{W^l}\log  \pi_l(H^{l-1}, H^l; W^l)].
	\end{equation}
	The proof is as follows:
	\begin{align}
		& \var[G\nabla_{W^l}\log  \pi_l(H^{l-1}, H^l; W^l)]  \label{eq:Q0}  \\
		=& \var[\ex[G\nabla_{W^l}\log  \pi_l(H^{l-1}, H^l; W^l)|S, A, G]] \nonumber \\ 
		&+ \ex[\var[G\nabla_{W^l}\log  \pi_l(H^{l-1}, H^l; W^l)|S, A, G]] \label{eq:Q1}\\	
		\geq& \var[\ex[G\nabla_{W^l}\log  \pi_l(H^{l-1}, H^l; W^l)|S, A, G]] \label{eq:Q2}\\
		=& \var[G \ex[\nabla_{W^l}\log  \pi_l(H^{l-1}, H^l; W^l)|S, A, G]] \label{eq:Q3}\\
		=& \var[G \ex[\nabla_{W^l}\log  \pi_l(H^{l-1}, H^l; W^l)|S, A]]. \label{eq:Q4}
	\end{align}
	
	(\ref{eq:Q0}) to (\ref{eq:Q1}) uses the law of total variance. \\
	(\ref{eq:Q1}) to (\ref{eq:Q2}) uses the fact that the second term must be non-negative. \\
	(\ref{eq:Q3}) to (\ref{eq:Q4}) uses the fact that $G$ is conditional independent with $H$ given $S$ and $A$.
	
	\section{MAP Propagation for Critic Networks}\label{sec:cn}
	
	Here we consider how to apply MAP propagation to a critic network. As the function of a critic network can be seen as approximating the scalar value $R_t + \gamma \hat{v}(S_{t+1})$, we consider how to learn a scalar regression task by MAP propagation in general. 
	
	A scalar regression task can be converted into a single-time-step MDP with the appropriate reward and $\mathbb{R}$ as the action set. For example, we can set the reward function to be $R(S, A) = -(A - A^*(S))^2$ (we dropped the subscript $t$ as it only has a single time step), with $A \in \mathbb{R}$ being the output of the network and $A^*(S) \in \mathbb{R}$ being the target output given input $S$. The maximization of rewards in this MDP is equivalent to the minimization of the L2 distance between the predicted value and the target value.
	
	But this conversion is inefficient since the information of the reward function is lost. In the following discussion, we restrict our attention to a network of stochastic units where all hidden layers and the output layer are normally distributed as defined in Section \ref{sec:n}. Let $\mu^L$ be the conditional mean of the output layer; that is, $\mu^L = g(H^{L-1}; W^{L})$. For this network, we propose an alternative learning rule that is similar to REINFORCE but with the return $G$ replaced by $(A^*(S)-\mu^L)/(A-\mu^L)$; that is, for $l \in \{1, 2, ..., L\}$:
	
	\begin{equation}
		W^l \leftarrow W^l + \alpha \frac{A^*(S)-\mu^L}{A-\mu^L} \nabla_{W^l}\log \pi_l(H^{l-1}, H^l; W^l). \label{eq:8a}
	\end{equation}
	
	It can be shown that after minimizing the energy function, the learning rule (\ref{eq:8a}) for the network is equivalent to gradient descent on the L2 error by backprop with the re-parameterization trick:

	\begin{theorem}\label{thm:3} 
		Let the policy be a multi-layer network of stochastic units with all hidden layers normally distributed as defined in Section \ref{sec:n} and the output layer has a single unit. There exists a re-parameterization of $H$ by ${Z}$ conditioned on $S$, denoted by $h(Z; W, S)$, such that for any $A^*: \mathcal{S} \rightarrow \mathbb{R}$, $l \in \{1, 2, ..., L\}$, $s \in \mathcal{S}$, $\hat{h}, \hat{z} \in \mathbb{R}^{n(1)} \times \mathbb{R}^{n(2)} \times ... \times \mathbb{R}^{n(L-1)}$ and $a \in \mathbb{R}$, if $\nabla_h E(\hat{h}; s, a) = 0$ and $\hat{z} = h^{-1}(\hat{h}; W, s)$, then
		\begin{align} 
			\frac{A^*(s)-\hat{\mu}^L}{a-\hat{\mu}^L} \nabla_{W^l}\log \pi_l(\hat{h}^{l-1}, \hat{h}^l; W^l) \propto - \nabla_{W^l} \left( A^*(s) -  \tilde{\mu}^L \right)^2,
		\end{align}
		where $\hat{\mu}^L := g^L(\hat{h}^{L-1}; W^{L})$ and $\tilde{\mu}^L := g^L(h^{L-1}(\hat{z}; W, s); W^{L})$.
	\end{theorem}

	%\begin{theorem}\label{thm:3} 
	%Let the policy be a multi-layer network of stochastic units with all hidden layers and output layer normally distributed as defined in Section \ref{sec:n}. 
	
	%Assume $\nabla_h E(\hat{H};S, A) = 0$. Then there exists a re-parameterization of $H$ by ${Z}$ conditioned on $S$, denoted by $h(Z; W, S)$, such that for $l \in \{1, 2, ..., L\}$,
	%\begin{align} 
	% \frac{A^*(S)-\hat{\mu}^L}{A-\hat{\mu}^L} \nabla_{W^l}\log \pi_l(\hat{H}^{l-1}, \hat{H}^l; W^l) \propto - \nabla_{W^l} \left( A^*(S) -  \tilde{\mu}^L \right)^2. 
	%\end{align}
	%where $\hat{Z} := h^{-1}(\hat{H}; W, S)$, $\hat{\mu}^L := g^L(\hat{H}^{L-1}; W^{L})$ and $\tilde{\mu}^L := g^L(h^{L-1}(\hat{Z}; W, S); W^{L})$.
	%\end{theorem}
	
	Therefore, we can apply the learning rule (\ref{eq:8a}) after minimizing the energy function by (\ref{eq:6}). The pseudo-code is the same as Algorithm \ref{alg:1} in Appendix \ref{sec:A0}, but with $G$ replaced by $(A^*(S)-\mu^L)/(A-\mu^L)$ in line 12.
	
	We then consider applying the above method to train a critic network, where the output of the network, $A_t \in \mathbb{R}$, is an estimation of the current value. In a critic network, the target output is $R_t + \gamma A_{t+1}$. However, a more stable estimate of target output is $R_t + \gamma \mu^L_{t+1}$ since the difference between $ A_{t+1}$ and $\mu^L_{t+1}$ is an independent Gaussian noise that can be removed. Therefore, we chose $A^*(S)$, the target output, as $R_t + \gamma \mu^L_{t+1}$ and TD error as  ${\delta}_t:= R_t + \gamma \mu^L_{t+1} - \mu^L_{t}$. Substituting back into (\ref{eq:8a}), the learning rule for the critic network becomes:
	\begin{equation}
		W^{l} \leftarrow W^{l} + \alpha \frac{{\delta}_t}{A_t-\mu^L_{t}} \nabla_{W^{l}} \log \pi_l(H^{l-1}_t, H^l_t; W^l), \label{eq:9}
	\end{equation}
	which is almost the same as the update rule for the actor network except the additional denominator $A_t-\mu^L_{t}$. The pseudo-code of training a critic network with eligibility trace using MAP propagation is the same as Algorithm \ref{alg:2} in Appendix \ref{sec:A0}, except (i) line 13 is replaced with ${\delta} \leftarrow \gamma \mu + R - \mu'$ where $\mu = g^L(H^{L-1};W^{L})$ and $\mu'$ is $\mu$ in the previous time step, and (ii) the gradient term in line 19 is multiplied by $(A-\mu)^{-1}$.
	
	Both the critic and the actor network can be trained together, and the TD error $\delta$ computed by the critic network can be passed to the actor network in line 13 of Algorithm \ref{alg:2}.
	
	\section{Details of Experiments} \label{sec:A4}
	
	In the multiplexer task, there are $k + 2^k$ binary inputs, where the first $k$ bits represent the address and the last $2^k$ bits represent the data, each of which is associated with an address. The output of the multiplexer is given by the value of data associated with the address. This is similar to the 2-bit multiplexer considered in \cite{barto1985learning}.
	
	We used Algorithm 1 in the multiplexer and the scalar regression experiment, and the hyperparameters can be found in Table \ref{table:2}. We used a different learning rate $\alpha$ for each layer of the network, and we denote the learning rate for the $l$\textsuperscript{th} layer to be $\alpha_l$. We denote the variance of the Gaussian distribution for the $l$\textsuperscript{th} layer to be $\sigma^2_l$.  The step size of hidden units when minimizing energy, $\alpha_h$, is selected to be 0.5 times the variance of the unit. For the learning rule in line 12 of the pseudo-code, we used Adam optimizer \cite{kingma2014adam} instead, with $\beta_1=0.9$ and $\beta_2=0.999$. We used batch update in both tasks, which means that we compute the parameter update for each sample in a batch, then we update the parameter using the average of these parameter updates. These hyperparameters are selected based on manual tuning to optimize the learning curve. We did the same manual tuning for the baseline models. 
	
	We used Algorithm 2 to train both the critic and the actor network in the experiments on RL tasks, and the hyperparameters can be found in Table \ref{table:3}. We did not use any batch update in our experiments, and we used a discount rate of $0.98$ for all tasks. The step size of hidden units when minimizing energy, $\alpha_h$, is selected to be 0.5 times the variance of the unit. We used Acrobot-v1, CartPole-v1, LunarLander-v2, and MountainCarContinuous-v0 in OpenAI's Gym for the implementation of the RL tasks.
	
	For the update rules in line 14 of Algorithm 2, we used Adam optimizer instead, with $\beta_1=0.9$ and $\beta_2=0.999$. Again, these hyperparameters are selected based on manual tuning to maximize the average return across all episodes. We did the same manual tuning for the baseline models.
	
	For the ANNs in the baseline models, the architecture is similar to the team of agents: 64 units on the first hidden layer and 32 units on the second hidden layer. If the output range is continuous, the output layer is a linear layer. If the output range is discrete, the output layer is a softmax layer. We used the softplus function as the activation function in the ANNs, which performs similarly to the ReLu function in our experiments.
	
	We annealed the learning rate linearly such that the learning rate is reduced to $\frac{1}{10}$ of the initial learning rate at $50000$ and $100000$ steps in CartPole and Acrobat respectively, and the learning rate remains unchanged afterward. We also annealed the learning rate linearly for the baseline models. We found that this can make the final performance more stable. For LunarLander and MountainCar, we did not anneal the learning rate. For MountainCar, we bound the reward by $\pm5$ to stabilize learning.
	
	\section{Experiments on MountainCar} \label{sec:A5}
	
	\begin{figure}
		\centering
		\begin{minipage}{.5\textwidth}
			\centering
			\includegraphics[width=.95\linewidth]{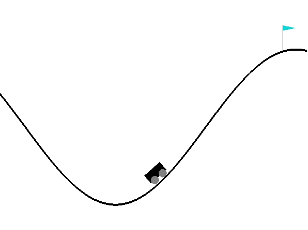}
			\captionof{figure}{Illustration of MountainCar.}
			\label{fig:A1}
		\end{minipage}%
		\begin{minipage}{.5\textwidth}
			\centering
			\includegraphics[width=\linewidth]{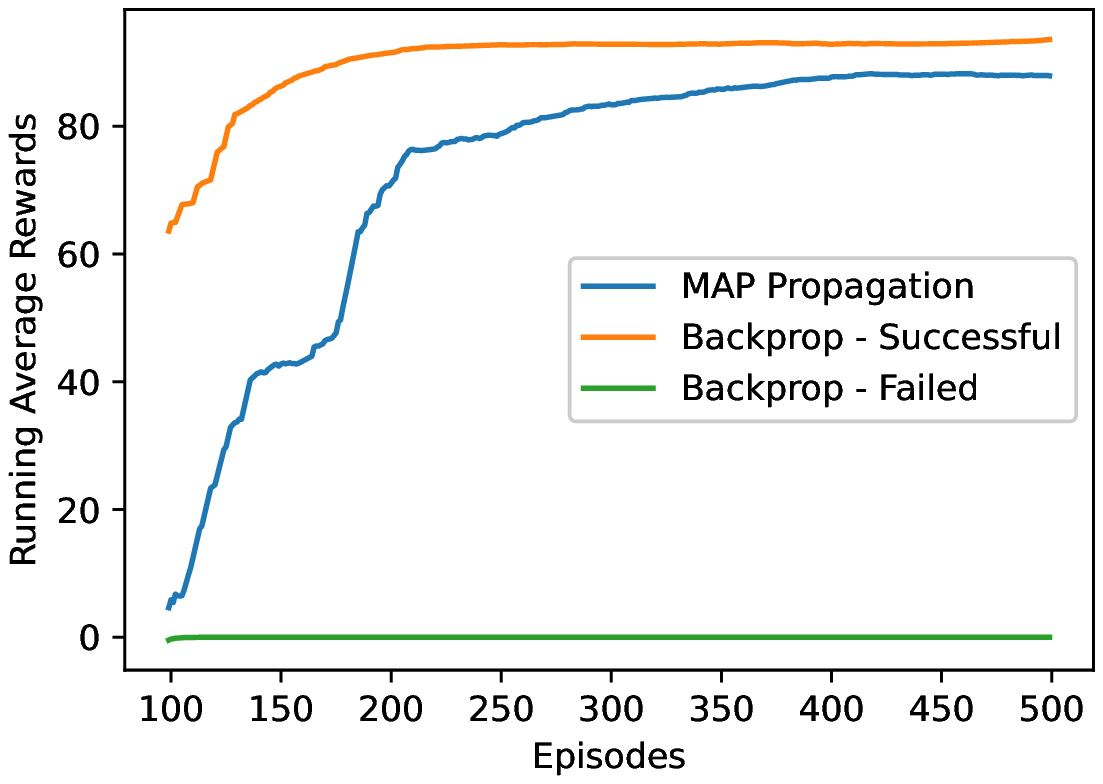}
			\captionof{figure}{Running average rewards over the last 100 episodes of the selected runs in MountainCar.}
			\label{fig:A2}
		\end{minipage}
	\end{figure}
	
	\begin{figure}
		\centering
		\includegraphics[width=1\textwidth]{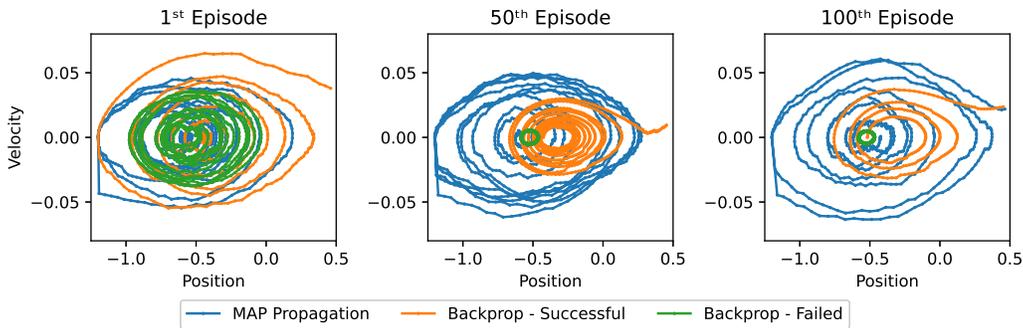}
		\caption{State trajectories of the 1\textsuperscript{st}, 50\textsuperscript{th} and 100\textsuperscript{th} episode of the selected runs. If the position is larger than 0.45, then the agent reaches the goal. Although the team of agents trained by MAP propagation did not reach the goal in both the 1\textsuperscript{st} and 50\textsuperscript{th} episode, the team was still exploring a large portion of the state space, which is in contrast to the failed baseline that stayed at the center. For the baseline, if it does not reach the goal in the first several episodes, it will be stuck in the local optima.
		}
		\label{fig:A3}  
	\end{figure}
	
	In the continuous version of MountainCar, the state is composed of two scalar values, which are the position and the velocity of the car, and the action is a scalar value corresponding to the force applied to the car. The goal is to reach the peak of the mountain on the right, where a large positive reward is given. However, to reach the peak, the agent has to first push the car to the left in order to gain enough momentum to overcome the slope. A small negative reward that is proportional to the force applied is given on every time step to encourage the agent to reach the goal with minimum force. An illustration of MountainCar is shown in Fig \ref{fig:A1}.
	
	One locally optimal policy is to apply zero force at every time step so that the car always stays at the bottom. In this way, the return is zero since no force is applied. We found that in many runs, the ANN trained by backprop was stuck in this locally optimal policy. However, in a few runs, the ANN can accidentally reach the goal in early episodes, which makes the ANN able to learn quickly and achieve an asymptotic reward of over +90, slightly higher than that of a team of agents trained by MAP propagation. The learning curve of a successful and a failed run of the agent trained by backprop is shown in Fig \ref{fig:A2}. 
	
	In contrast, for teams of agents trained by MAP propagation, the teams in all runs can learn a policy that reaches the goal successfully. However, this comes at the expense of slower learning and a slightly worse asymptotic performance, as seen from the learning curve of a typical run of the team of agents trained by MAP propagation shown in Fig \ref{fig:A2}. This is likely due to the larger degree of exploration in MAP propagation, which can be illustrated by the state trajectories shown in Fig \ref{fig:A3}. 
	
	We used the same variance for the output unit in both MAP propagation and the baseline models. We found that even using a larger variance or adding entropy regularization cannot prevent the baseline models from being stuck in the local optima. This suggests that MAP propagation allows more sophisticated exploration instead of merely more exploration. In a team of agents, each agent in the team is exploring its own action space, thus allowing exploration in different levels of the hierarchy. This may explain the difference in exploration behavior observed in a team of agents trained by MAP propagation compared to an ANN trained by backprop.
	
	%From this experiment, we see that the learning dynamic of a team of agents trained by MAP propagation is very different from that of an ANN trained by backprop. A team of agents trained by MAP propagation can prevent local optima better but at the expense of slower learning and worse asymptotic performance. This may be explained by the more sophisticated exploration of MAP propagation due to its exploration in different levels of the hierarchy.
	
	\section{Biological Plausibility of MAP Propagation} \label{sec:A7}
	
	As discussed in the paper, backprop has three major problems with biological plausibility due to the requirement of 1. non-local information in the learning rule, 2. precise coordination between the feedforward and feedback phase, and 3. symmetry of feedforward and feedback weights. However, REINFORCE does not have any of these issues. Other than the global reinforcement signal, the learning rule of REINFORCE does not depend on non-local information. Also, since REINFORCE does not require any feedback connections, the second and third issues do not exist for REINFORCE. We refer readers to chapter 15 of \cite{sutton2018reinforcement} for a review and discussion of the connection between REINFORCE and neuroscience.
	
	Compared to backprop, REINFORCE is more consistent with biologically-observed forms of synaptic plasticity. When applied to a Bernoulli-logistic unit, REINFORCE gives a three-factor learning rule which depends on a reinforcement signal, input, and output of the unit. This is similar to R-STDP observed biologically, which depends on a neuromodulatory input, presynaptic spikes, and postsynaptic spikes. It has been proposed that dopamine, a neurotransmitter, plays the role of neuromodulatory input in R-STDP and corresponds to the TD error from RL. 
	
	Despite the elegance of REINFORCE, its learning speed is far lower than backprop and scales poorly with the number of units in the network since only a scalar feedback is used to assign credit to all units in the network. It can be argued that learning speed may not be the issue for the biological plausibility of REINFORCE, given that evolution already equips the brain with prior knowledge, and the brain can learn from experience accumulated over the entire lifetime. However, the learning speed of REINFORCE may not explain many remarkable learning behaviors of humans, such as mastering Go, despite the fact that the ability to play Go likely does not come from prior knowledge shaped by evolution. Given billions of neurons in the brain, it is likely that the brain employs some forms of structural credit assignment to speed up learning.
	
	MAP propagation presents one possible solution for structural credit assignment. Essentially, MAP propagation is equivalent to applying REINFORCE after minimizing the energy function. The idea of minimizing the energy function is to nudge the values of hidden units towards those that are more consistent with the values of units on the first and last layer, i.e.\ the state and the action. In this process, feedback connections are required to drive the value of units, so as to propagate information from the layers above. 
	
	Although the purpose of feedback connections in the brain is still not completely understood, there has been evidence showing that feedback connections can drive the activity of neurons \cite{khayat2009time, roelfsema2010perceptual, roelfsema2018control}. For example, in the visual system, the activity of neurons that is responsible for the selected action will be enhanced by feedback connection \cite{khayat2009time}. This is analogous to the updates of hidden units in MAP propagation. The general idea that feedback connection drives the activity of units in lower layers to facilitate local learning rules underlies many proposals for biological learning and machine learning algorithms \cite{lillicrap2020backpropagation}. This idea is also fundamental to the NGRAD hypothesis \cite{lillicrap2020backpropagation}, which will be discussed next.
	
	NGRAD hypothesis is a recently proposed hypothesis that unifies many biologically plausible alternatives to backprop with local learning rules \cite{o1996biologically,  bengio2015towards, whittington2017approximation, hinton2002training, scellier2017equilibrium, Hinton1988Neural,  ackley1985learning, xie2003equivalence, bengio2014auto}. It hypothesizes that the cortex uses the differences in activity states to drive learning, and the induced differences are brought by the nudging of lower-level activities towards those values that are more consistent with the high-level activities. In this way, local learning rules can yield an approximation to backprop without storing units' values and error signals at the same time.
	
	MAP propagation fits well into the NGRAD hypothesis. When normally distributed hidden units are settled to the minima of the energy function, REINFORCE, a local learning rule except for the global reinforcement signal, yields the same parameter update given by backprop. Therefore, MAP propagation can be seen as an approximation of backprop by changing the values of hidden units. However, MAP propagation has major differences with many other algorithms based on the NGRAD hypothesis (NGRAD algorithms). First, most NGRAD algorithms require storage of past units' values (e.g.\ in target propagation \cite{bengio2014auto}, the unit has to store its past value to compute the reconstruction error) or separate phases of learning (e.g.\ the positive and negative phase in contrastive divergence \cite{hinton2002training}), but MAP propagation requires neither of them. Second, most NGRAD algorithms require precise coordination between feedforward and feedback connections (e.g.\ in target propagation, the unit has to coordinate between computing the reconstruction error and adding it to the uncorrelated target). In contrast, the updates for all layers can be done in parallel without any coordination between feedforward and feedback connections in MAP propagation. Third, MAP propagation is derived based on RL, while NGRAD algorithms are derived based on either supervised or unsupervised learning. Given the observation of R-STDP in biological systems and the correspondence between R-STDP and REINFORCE, MAP propagation presents a new paradigm of explaining biological learning in NGRAD algorithms. However, a major limitation of MAP propagation is that it requires a different value to be propagated through feedforward and backward connections.
	
	To see this, we will closely examine the update rule (\ref{eq:6}) for minimizing energy function in MAP propagation. Assuming all units are normally distributed with a fixed variance; i.e.\ $\pi_l(H^{l-1},  \cdot; W^l) = N(f(W^{l}H^{l-1}), \sigma^2)$ for $l \in \{1, 2, ..., L\}$ and $f$ is a non-linear activation function, then the update rule (\ref{eq:6}) and the learning rule (\ref{eq:1}) becomes \footnote{We ignore the subscript $t$ here since it does not affect our discussion, and $\odot$ denotes element-wise multiplication.}:
	\begin{align}
		&\Delta H^l = \frac{1}{\sigma^2}\left(-e^l +  (W^{l+1})^T (e^{l+1} \odot \delta^{l+1})\right),\label{eq:N0} \\
		&\Delta W^l = \frac{1}{\sigma^2}\left( G \cdot e^l (H^{L-1})^T\right),\\
		\text{where } &\mu^l = f(W^{l}H^{l-1}), \; \delta^{l} =  f'(W^{l}H^{l-1}), \; e^l = H^l - \mu^l \text{ for } l \in \{1, 2, ..., L\}.  \nonumber
	\end{align}
	
\begin{figure}
		\centering
		\includegraphics[width=0.65
		\textwidth]{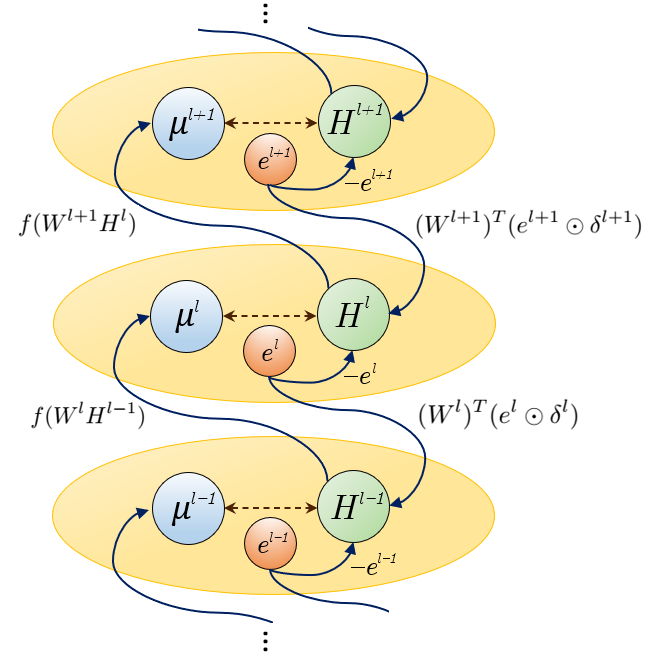}
		\caption{Illustration of minimizing the energy function in MAP propagation. The mean value of units on the layer $l$, denoted by $\mu^l$, is computed as a function of the value of units on the previous layer, denoted by $H^{l-1}$. The difference between the current and the mean value of units, denoted by $e^l$, is then used to drive the value of units on the same layer ($H^l$) and the value of units on the previous layer ($H^{l-1}$).}
		\label{fig:A4}  
	\end{figure}

	Both the update rule (\ref{eq:6}) and the learning rule (\ref{eq:1}) are local and can be applied to all hidden layers in parallel. There are two components in the update rule: i. the feedforward signal $-e^l$ and ii. the feedback signal $(W^{l+1})^T (e^{l+1} \odot \delta^{l+1})$. The feedforward signal nudges the value of the unit, $H^l$, closer to the mean value of the unit, $\mu^l$, which only depends on feedforward signals. However, it is not yet clear how the feedback signal can be implemented with biological systems. First, it requires information to be propagated through the feedback weight $(W^{l+1})^T$ that is symmetric of the feedforward weight in the next layer, which may not be biologically plausible. Nonetheless, recent work has shown that symmetric weights may not be necessary for backprop due to the phenomenon of `feedback alignment' \cite{lillicrap2014random, lillicrap2016random, liao2016important}, and similar phenomenons may also exist for MAP propagation. Second, the information to be propagated backward is $e^{l+1} \odot \delta^{l+1}$, which is different from the information to be propagated forward ($H^l$). An illustration of this is shown in Fig \ref{fig:A4}.

	The issue of propagating two different values also exists for backprop since error signals, instead of units' values, are propagated backward in backprop. However, in backprop, the backpropagated error signals have to be stored separately from the units' values, so as to compute the next error signals to be passed to the lower layers. In contrast, the feedback signal in MAP propagation is only used to nudge the value of units and does not need to be stored separately. In other words, backprop requires non-local information in the computation of feedback signal, but not MAP propagation.
	
	It is not yet clear how a neuron can propagate two different values at the same time, even if both values are locally available. But there are many possible solutions to avoid propagating different values in MAP propagation. For instance, since the feedback signal $(W^{l+1})^T (e^{l+1} \odot \delta^{l+1})$ can be expressed as a function of $H^l$ and $H^{l+1}$, it might be possible to approximate this term based on feedback connections (for $H^{l+1}$) and recurrent connections (for $H^{l}$), and learn the weights in these connections, such that all units are propagating the same value. Another possible solution is to minimize the energy function by hill-climbing methods instead of gradient ascent, such that only the scalar energy, instead of feedback connection, is required to guide the minimization of the energy function. Further work can be done on these possible solutions.
	
	Despite the limitations of MAP propagation, we argue that MAP propagation is more biologically plausible than backprop. The two major limitations of MAP propagation also exist in backprop, but MAP propagation does not require non-local information in both the learning rule or the computation of feedback signals. Also, the update of all layers can be done in parallel in MAP propagation, removing the requirement of precise coordination between feedforward and feedback connections that is required in backprop.
	
	\clearpage
	\begin{table}
		\caption{Hyperparameters used in multiplexer and scalar regression experiments.}
		\label{table:2}
		\begin{center}	
			\begin{tabular}{lcccc}
				\toprule[0.1ex]
				& Multiplexer & Scalar Regression \\
				\midrule
				Batch Size & 128 & 128 \\
				$N$ & 20 & 20  \\	    
				$\alpha_1$ & 4e-2 & 6e-2\\	 
				$\alpha_2$ & 4e-5 & 6e-5\\	 
				$\alpha_3$ & 4e-6 & 6e-6\\	 	
				$\sigma^2_1$ & 0.3 & 0.0075\\
				$\sigma^2_2$ & 1 & 0.025\\
				$\sigma^2_3$ & n.a. & 0.025\\		
				$T$ & 1 & n.a. \\		
				\bottomrule[0.25ex]
			\end{tabular}
		\end{center}
	\end{table}
	
	\begin{table}
		\caption{Hyperparameters used in Acrobat, Cartpole,  Lunarlander and MountainCar experiments.}
		\label{table:3}
		\begin{center}
			\begin{tabular}{lcccccccc}
				\toprule[0.1ex]
				& \multicolumn{2}{c}{Acrobat} & \multicolumn{2}{c}{CartPole} & \multicolumn{2}{c}{LunarLander} & \multicolumn{2}{c}{MountainCar}\\
				\cmidrule(r){2-3}  \cmidrule(r){4-5} \cmidrule(r){6-7} \cmidrule(r){8-9}
				& Critic & Actor & Critic  & Actor  & Critic  & Actor & Critic  & Actor \\
				\midrule
				$N$ & 20 & 20 & 20 & 20 & 20 & 20 & 20 & 20\\	    		
				$\alpha_1$ & 2e-2 & 1e-2 & 2e-2 & 1e-2 & 1e-2 & 4e-3 & 1e-2 & 4e-3\\	 
				$\alpha_2$ & 2e-5 & 1e-5 & 2e-5 & 1e-5 & 1e-5 & 4e-6 & 1e-5 & 4e-6\\	 
				$\alpha_3$ & 2e-6 & 1e-6 & 2e-6 & 1e-6 & 1e-6 & 4e-7 & 1e-6 & 4e-7\\	 
				$\sigma^2_1$ & 0.06 & 0.03 & 0.03 & 0.03 & 0.003 & 0.06 & 0.003 & 0.03 \\		
				$\sigma^2_2$ & 0.2 & 0.1 & 0.1 & 0.1 & 0.01 & 0.2 & 0.01 & 0.1 \\		
				$\sigma^2_3$ & 0.2 & n.a. & 0.1 & n.a. & 0.01 & n.a. &  0.05 & 0.5 \\		
				$T$ & n.a. & 4 & n.a. & 2 & n.a. & 8 & n.a. & n.a. \\		
				$\lambda$ & .97 & .97 & .95 & .95 & .97 & .97 & .97 & .97 \\	
				\bottomrule[0.25ex]
			\end{tabular}
		\end{center}
	\end{table}
	
	%Errata: Theorem 2: "Let .. to be" > "Let ... be"
	%Footnote 2: $\hat{h}_t$ > $\hat{h}$ 
	
	\clearpage

\end{document}